\def\arxivversion{}
\documentclass[letterpaper]{article} %

\ifdefined\arxivversion
\usepackage[preprint]{aaai2027}
\else
\usepackage[submission]{aaai2027}  %
\fi
\usepackage[hyphens]{url}  %
\usepackage{graphicx} %
\usepackage{natbib}  %
\usepackage{caption} %
\usepackage[utf8]{inputenc}
\usepackage{amsmath}
\usepackage{amssymb}        %
\usepackage{booktabs}       %
\usepackage{tabularx}       %
\usepackage{tikz}           %
\usetikzlibrary{arrows.meta, positioning, shapes.geometric, calc, backgrounds, fit, decorations.markings, patterns, shapes.symbols, shadows}
\usepackage[most]{tcolorbox}
\usepackage{colortbl}       %

\makeatletter
\newcommand{\apstroke}{%
  \expandafter\def\expandafter\tcb@upperdraw@begin@hook\expandafter{%
    \tcb@upperdraw@begin@hook\pdfliteral page{2 Tr 0.25 w}}%
  \expandafter\def\expandafter\tcb@upperdraw@end@hook\expandafter{%
    \tcb@upperdraw@end@hook\pdfliteral page{0 Tr 1 w}}}
\makeatother
\newcommand{\boxfont}{\footnotesize}
\tcbset{apbox/.style={breakable, code=\apstroke, boxrule=0.4pt, arc=1.5pt, boxsep=1pt,
  left=3pt, right=3pt, top=1.5pt, bottom=1.5pt, toptitle=0.5pt, bottomtitle=0.5pt,
  before skip=4pt, after skip=4pt, colupper=black, fontupper=\boxfont,
  fonttitle=\boxfont\bfseries}}
\newtcolorbox{promptbox}[1]{apbox, colback=white, colframe=black, coltitle=black,
  colbacktitle=black!10, title={#1}}
\newtcolorbox{personabox}[1]{apbox, colback=white, colframe=black, coltitle=black,
  colbacktitle=black!10, title={#1}, before skip=2pt, after skip=2pt}
\newtcolorbox{instrbox}[1]{apbox, colback=white, colframe=black, coltitle=black,
  colbacktitle=black!10, title={#1}}
\newcommand{\hd}[1]{\begin{tabular}[b]{@{}c@{}}#1\end{tabular}}
\newcommand{\pt}[1]{{\ttfamily\footnotesize #1}}
\newcommand{\ci}[2]{\mbox{[#1,\,#2]}}
\newcommand{\appref}[2]{\ifdefined\mainonly #2\else\ref{#1}\fi}

\definecolor{aenavy}{HTML}{1D4E89}   %
\definecolor{aebrown}{HTML}{9A5B32}  %
\definecolor{aeteal}{HTML}{2A9D8F}   %
\definecolor{aegray}{HTML}{8A8A86}
\definecolor{aegood}{HTML}{0CA30C}   %
\definecolor{aebad}{HTML}{D03B3B}    %
\definecolor{aewarn}{HTML}{FAB219}   %
\colorlet{aesyn}{aenavy}             %
\colorlet{aereal}{aebrown}           %
\colorlet{aerowmean}{aesyn!10}
\colorlet{aerowalt}{black!5}

\newcommand{\qwenmark}{\raisebox{-0.75ex}{\includegraphics[height=7pt]{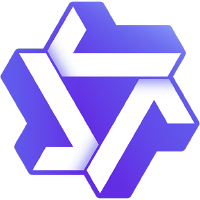}}}
\newcommand{\fluxmark}{\raisebox{-0.55ex}{\includegraphics[height=6pt]{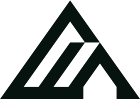}}}
\newcommand{\dinomark}{\raisebox{-0.45ex}{\includegraphics[height=5pt]{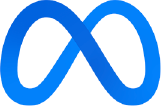}}}

\title{Agreement Is Not Fidelity: Auditing VLM Personas as a Synthetic Visual Audience}
\author{
    Sajjad Pakdamansavoji\textsuperscript{\rm 1}\corresponding,
    Amin Fadaeinejad\textsuperscript{\rm 1}
}
\affiliations{
    \textsuperscript{\rm 1}Forall Labs\\
    \{sj.pakdaman.edu, aminfadaeinejad.edu\}@gmail.com
    \ifdefined\arxivversion\\\url{https://forall-labs.github.io/AutoPolish}\fi
}

\begin{document}

\maketitle

\begin{abstract}
Synthetic audiences, in which one vision-language model (VLM) answers many times as viewers described by short profiles, are proposed as stand-ins for people who judge and refine images. Agreement with a human panel does not show that the personas are faithful, because calibration, the composition of the scored panel, fixed group offsets, prompt wording and selection can each produce it. We audit a frozen Qwen2-VL-7B-Instruct audience whose cards are written from real raters' recorded profiles, rating photographs and paintings from three datasets with greedy decoding, and pair each kind of agreement with a comparator that could explain it. After identical held-out calibration on human ratings, a persona-free prompt predicts each photograph's mean rating more accurately than the persona prompt. The persona prompt is more accurate on paintings, where the prompts also differ, so this gain cannot be assigned to the cards alone. Across disjoint halves of each image's raters, predicted gaps between interest-in-art groups agree with the real gaps at 0.23 (uncentered correlation), but only at 0.03 once each group pair's average gap is removed, and on EVA the predicted gaps run opposite to the real ones. Attributes remain decodable from generated comments after card words are masked (AUC 0.58 to 0.70), even where ratings invert the real gap. As an image-editing critic, the audience shows no detected gain over a fixed instruction under the loop's own scorer (0.006, 95\% interval $-0.020$ to 0.034) at 110 extra model calls per image. Aggregate accuracy, group gaps, persona language and editing gains thus support different kinds of validity, and we summarize the comparators as a checklist for population simulations.
\end{abstract}

\section{Introduction}

\begin{figure*}[t]
  \centering
  \begin{tikzpicture}[
    font=\scriptsize,
    person/.pic={
      \path[pic actions] (0,0.135) circle (0.075);
      \path[pic actions] (-0.13,-0.11) .. controls (-0.13,0.03) and (-0.08,0.045) .. (0,0.045)
        .. controls (0.08,0.045) and (0.13,0.03) .. (0.13,-0.11) -- cycle;},
    mask/.pic={
      \path[pic actions, even odd rule]
        (-0.22,0.05) .. controls (-0.16,0.12) and (-0.05,0.11) .. (0,0.065)
        .. controls (0.05,0.11) and (0.16,0.12) .. (0.22,0.05)
        .. controls (0.21,-0.05) and (0.11,-0.10) .. (0.045,-0.055)
        .. controls (0.02,-0.025) and (-0.02,-0.025) .. (-0.045,-0.055)
        .. controls (-0.11,-0.10) and (-0.21,-0.05) .. cycle
        (-0.095,0.02) ellipse (0.055 and 0.032) (0.095,0.02) ellipse (0.055 and 0.032);},
    robot/.pic={
      \draw[pic actions, line width=0.6pt, rounded corners=1.5pt] (-0.19,-0.15) rectangle (0.19,0.13);
      \draw[pic actions, line width=0.6pt] (0,0.13) -- (0,0.2);
      \draw[pic actions, line width=0.6pt] (0,0.225) circle (0.025);
      \draw[pic actions, line width=0.6pt] (-0.235,-0.06) rectangle (-0.19,0.04) (0.19,-0.06) rectangle (0.235,0.04);
      \draw[pic actions, line width=0.6pt, line cap=round] (-0.07,-0.095) -- (0.07,-0.095);},
    tile/.style={rounded corners=5pt, draw=black!35, line width=0.7pt},
    ttl/.style={font=\footnotesize\bfseries, anchor=north west, inner sep=0},
    sub/.style={font=\scriptsize, text=black!60, anchor=north west, inner sep=0},
    box/.style={rounded corners=3pt, line width=0.6pt},
    chip/.style={rounded corners=2.5pt, fill=aereal!10, draw=aereal!45, line width=0.3pt,
                 inner sep=1.3pt, font=\tiny, minimum height=0.27cm, text width=1.8cm, align=center},
    think/.style={cloud, cloud puffs=11, aspect=1.9, draw=#1, fill=#1!8, line width=0.5pt,
                 inner sep=0.3pt, font=\scriptsize\bfseries, text=#1},
    real/.style={circle, fill=aereal, minimum size=5.0mm, inner sep=0pt, text=white,
                 font=\scriptsize\bfseries},
    sim/.style={circle, fill=aesyn, minimum size=5.0mm, inner sep=0pt, text=white,
                font=\scriptsize\bfseries},
    reallink/.style={-{Latex[length=1.5mm]}, draw=aereal, line width=0.6pt},
    simlink/.style={-{Latex[length=1.5mm]}, draw=aesyn, line width=0.6pt},
    vlabel/.style={anchor=north west, inner sep=0, font=\tiny\bfseries},
    vnote/.style={anchor=north west, inner sep=0, font=\tiny, text=black!65, align=left,
                  text width=#1},
    no/.style={font=\footnotesize\bfseries, text=aebad, inner sep=0},
    yes/.style={font=\footnotesize\bfseries, text=aegood, inner sep=0},
    stage/.style={rounded corners=2.5pt, fill=white, draw=black!50, line width=0.45pt,
                 align=center, inner sep=2pt, font=\tiny},
    ringin/.style={line width=1.0pt, postaction={decorate}, decoration={markings,
                 mark=at position 0.1180 with {\arrow{Latex[length=2mm]}},
                 mark=at position 0.4156 with {\arrow{Latex[length=2mm]}},
                 mark=at position 0.6179 with {\arrow{Latex[length=2mm]}},
                 mark=at position 0.9155 with {\arrow{Latex[length=2mm]}}}},
    ringout/.style={line width=1.0pt, postaction={decorate}, decoration={markings,
                 mark=at position 0.1159 with {\arrow{Latex[length=2mm]}},
                 mark=at position 0.4123 with {\arrow{Latex[length=2mm]}},
                 mark=at position 0.6043 with {\arrow{Latex[length=2mm]}},
                 mark=at position 0.9238 with {\arrow{Latex[length=2mm]}}}},
  ]
  \newcommand{\audlogo}[2]{%
    \begin{scope}[shift={#1}, scale=#2]
      \fill[aesyn!10] (0,0) circle (0.42); \draw[aesyn, line width=0.6pt] (0,0) circle (0.42);
      \pic[fill=aesyn!50, scale=0.62] at (-0.15,0.12) {mask};
      \pic[fill=aesyn!50, scale=0.62] at (0.15,0.12) {mask};
      \pic[fill=aesyn, scale=0.78] at (0,-0.1) {mask};
    \end{scope}}
  \newcommand{\reallogo}[2]{%
    \begin{scope}[shift={#1}, scale=#2]
      \fill[aereal!10] (0,0) circle (0.42); \draw[aereal, line width=0.6pt] (0,0) circle (0.42);
      \pic[fill=aereal!50, scale=0.82] at (-0.17,0.05) {person};
      \pic[fill=aereal!50, scale=0.82] at (0.17,0.05) {person};
      \pic[fill=aereal, draw=aereal!10, line width=0.5pt, scale=1.0] at (0,-0.06) {person};
    \end{scope}}
  \newcommand{\tilehead}[3]{%
    \node[ttl] (tl) at (0.2,-0.15) {#1};
    \node[ttl] (tt) at ($(tl.north east)+(0.1,0)$) {#2};
    \node[sub] at ($(tt.south west)+(0,-0.07)$) {#3};}
  \newcommand{\finding}[5]{%
    \ifnum\pdfstrcmp{#2}{part}=0
      \fill[aewarn] ($#1+(0.05,-0.13)$) ++(90:0.085) arc (90:270:0.085) -- cycle;
      \draw[black!65, line width=0.5pt] ($#1+(0.05,-0.13)$) circle (0.085);
    \else
      \node[#2] at ($#1+(0.05,-0.13)$) {\ifnum\pdfstrcmp{#2}{no}=0 $\boldsymbol{\times}$\else $\checkmark$\fi};
    \fi
    \node[vlabel] at ($#1+(0.22,0)$) {#3};
    \node[vnote=#5] at ($#1+(0.22,-0.19)$) {#4};}
  \newcommand{\scorebars}[3]{%
    \begin{scope}[shift={#1}]
      \foreach \v/\n/\s in {#3}
        \path[#2=\s] (0.13*\v,0) rectangle +(0.1,0.028*\n);
      \draw[black!55, line width=0.4pt] (-0.03,0) -- (1.43,0);
      \foreach \v in {0,5,10}
        \node[font=\tiny, text=black!55, anchor=north, inner sep=1pt] at (0.13*\v+0.05,0) {\v};
    \end{scope}}

  \begin{scope}
  \draw[tile] (0,0) rectangle (5.4,-5.3);
  \tilehead{(a)}{VLM persona vs.\ real rater}{does a profile predict one person's ratings?}

  \node[inner sep=0, draw=black, line width=0.6pt] (ph) at (2.7,-1.62) {\includegraphics[width=1.05cm]{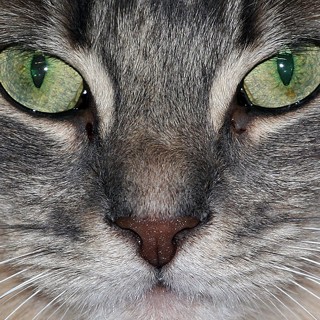}};
  \pic[draw=aesyn, fill=white, scale=2.0] at (0.98,-1.62) {robot};
  \pic[fill=aesyn, scale=1.75] at (0.98,-1.58) {mask};
  \node[font=\scriptsize\bfseries, anchor=north, text=aesyn] at (0.98,-1.95) {VLM persona};
  \draw[simlink] (ph.west) -- (1.51,-1.62);
  \node[think=aesyn] at (1.55,-1.1) {8/10};
  \fill[aesyn!70] (1.21,-1.31) circle (0.032) (1.12,-1.37) circle (0.022);
  \pic[fill=aereal, scale=2.0] at (4.62,-1.62) {person};
  \node[font=\scriptsize\bfseries, text=aereal, anchor=north] at (4.62,-1.95) {real rater};
  \draw[reallink] (ph.east) -- (4.25,-1.62);
  \node[think=aereal] at (3.98,-1.1) {1/10};
  \fill[aereal!70] (4.34,-1.33) circle (0.032) (4.43,-1.39) circle (0.022);

  \draw[box, draw=aesyn, fill=aesyn!5] (0.2,-2.42) rectangle (2.4,-4.27);
  \pic[fill=aesyn, scale=1.05] at (0.47,-2.62) {mask};
  \node[anchor=west, font=\scriptsize\bfseries, text=aesyn, inner sep=0] at (0.72,-2.62) {persona card};
  \node[anchor=north west, align=left, text width=1.98cm, font=\tiny, inner sep=0] at (0.32,-2.92)
    {``a male person aged 28 from Mainland China; is a beginner at photography.''};
  \draw[simlink] (2.95,-3.46) -- (2.45,-3.46);
  \draw[box, draw=aereal, fill=aereal!4] (3.0,-2.42) rectangle (5.2,-4.27);
  \pic[fill=aereal, scale=0.85] at (3.22,-2.62) {person};
  \node[anchor=west, font=\scriptsize\bfseries, text=aereal, inner sep=0] at (3.4,-2.62) {dataset record};
  \foreach \y/\t in {-2.98/{male}, -3.3/{born 1992}, -3.62/{Mainland China},
                     -3.94/{photo: beginner}}
    \node[chip] at (4.1,\y) {\t};

  \draw[black!20, line width=0.4pt] (0.2,-4.45) -- (5.2,-4.45);
  \finding{(0.25,-4.57)}{no}{Profiles do not predict stable rater offsets}
    {Tested profile regressions do worse than a constant on held-out raters' offsets.}{4.95cm}
  \end{scope}

  \begin{scope}[xshift=5.6cm]
  \draw[tile] (0,0) rectangle (6.3,-5.3);
  \tilehead{(b)}{Synthetic audience vs.\ real panel}{one simulated rating for every real one}

  \audlogo{(1.1,-1.3)}{1.0}
  \node[font=\scriptsize\bfseries, text=aesyn, anchor=north] at (1.1,-1.76) {synthetic audience};
  \reallogo{(5.2,-1.3)}{1.0}
  \node[font=\scriptsize\bfseries, text=aereal, anchor=north] at (5.2,-1.76) {real panel};
  \foreach \i/\c in {0/aesyn!55, 1/aesyn!30, 2/aesyn!10}
    \draw[rounded corners=2pt, fill=\c, draw=aesyn, line width=0.45pt]
      ($(0.55,-2.2)+(0.09*\i,-0.09*\i)$) rectangle +(1.0,-0.58);
  \pic[fill=aesyn, scale=1.0] at (1.23,-2.67) {mask};
  \foreach \i/\c in {0/aereal!55, 1/aereal!30, 2/aereal!10}
    \draw[rounded corners=2pt, fill=\c, draw=aereal, line width=0.45pt]
      ($(4.65,-2.2)+(0.09*\i,-0.09*\i)$) rectangle +(1.0,-0.58);
  \pic[fill=aereal, scale=1.2] at (5.33,-2.67) {person};
  \node[inner sep=0, draw=black, line width=0.6pt] (pb) at (3.15,-1.95) {\includegraphics[width=1.15cm]{figs_aaai/ov_cat.jpg}};
  \draw[simlink, rounded corners=3pt] (3.0,-2.55) |- (1.8,-2.72);
  \draw[reallink, rounded corners=3pt] (3.3,-2.55) |- (4.6,-2.72);
  \scorebars{(0.42,-3.72)}{fill=aesyn, fill opacity}
    {6/2/0.35, 7/5/0.6, 8/17/1, 9/6/0.75, 10/2/0.45}
  \scorebars{(4.5,-3.72)}{fill=aereal, fill opacity}
    {0/3/0.4, 1/6/0.65, 2/8/0.85, 3/8/1, 4/6/0.8, 5/4/0.6, 6/3/0.45, 7/2/0.35, 8/1/0.3}
  \draw[simlink] (1.23,-3.0) -- (1.23,-3.52);
  \draw[reallink] (5.33,-3.0) -- (5.33,-3.52);
  \draw[simlink] (0.8,-3.75) |- (2.93,-4.05);
  \draw[reallink] (5.55,-3.75) |- (3.37,-4.05);
  \node[circle, draw=black!55, line width=0.5pt, fill=white, inner sep=0.5pt, font=\scriptsize]
    at (3.15,-4.05) {$\overset{?}{=}$};

  \draw[black!20, line width=0.4pt] (0.2,-4.45) -- (6.1,-4.45);
  \finding{(0.2,-4.57)}{no}{Image means}{persona-free\\wins on photos}{1.6cm}
  \finding{(2.3,-4.57)}{no}{Group gaps}{mostly an offset,\\reversed on EVA}{1.6cm}
  \finding{(4.35,-4.57)}{part}{Persona voice}{in words,\\not in ratings}{1.6cm}
  \end{scope}

  \begin{scope}[xshift=12.1cm]
  \draw[tile] (0,0) rectangle (5.64,-5.3);
  \tilehead{(c)}{Audience as an editing critic}{critique, edit, keep the best, repeat}

  \coordinate (O) at (2.82,-2.5);
  \draw[ringout, aegray] ($(O)+(180:1.13)$) arc (180:-180:1.13);
  \draw[ringin, aesyn]  ($(O)+(180:0.95)$) arc (180:-180:0.95);
  \node[stage, minimum width=1.25cm, minimum height=0.56cm] (cr) at ($(O)+(0,1.04)$) {};
  \audlogo{($(cr.center)+(-0.34,0)$)}{0.52}
  \node[font=\tiny, text=aegray, align=center, inner sep=0] at ($(cr.center)+(0.34,0)$) {fixed\\[-2pt]instr.};
  \node[font=\small, text=black!60] at (cr.center) {/};
  \node[font=\tiny\bfseries, anchor=south, inner sep=1pt] at (cr.north) {critic};
  \node[stage, minimum width=0.66cm, minimum height=0.66cm, inner sep=0.5pt] (fx) at ($(O)+(1.04,0)$) {\includegraphics[height=5pt]{figs/flux_logo.png}\\[-1pt]FLUX\\[-1pt]edits};
  \node[stage, minimum width=1.25cm, minimum height=0.56cm, inner sep=0.5pt] (sc) at ($(O)+(0,-1.04)$) {\raisebox{-0.3ex}{\includegraphics[height=6pt]{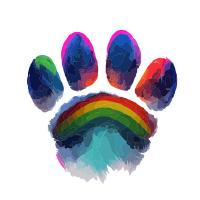}}\,LAION scorer};
  \node[inner sep=0, draw=black, line width=0.5pt] (im) at ($(O)+(-1.04,0)$)
    {\includegraphics[width=0.66cm]{figs_aaai/ov_cat.jpg}};
  \node[font=\tiny\itshape, text=black, align=center] at ($(O)+(0,0.12)$)
    {``remove green\\[-1pt]eyes and make\\[-1pt]cat's eyes\\[-1pt]normal''};
  \node[font=\tiny, text=black!55] at ($(O)+(0,-0.55)$) {10 rounds};

  \node[inner sep=0, draw=aesyn, line width=1.3pt] (endb) at (0.72,-2.75)
    {\includegraphics[width=1.05cm]{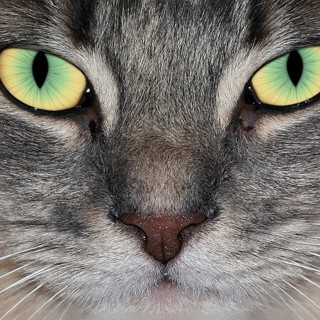}};
  \node[font=\tiny\bfseries, text=aesyn, anchor=south, inner sep=1.5pt, align=center] at (endb.north)
    {audience {\color{aebad}\footnotesize$\boldsymbol{\times}$}\\[-1pt]{\mdseries(kept at round 2,}\\[-1pt]{\mdseries later replaced)}};
  \node[inner sep=0, draw=aegray, line width=1.3pt] (endg) at (4.92,-2.75)
    {\includegraphics[width=1.05cm]{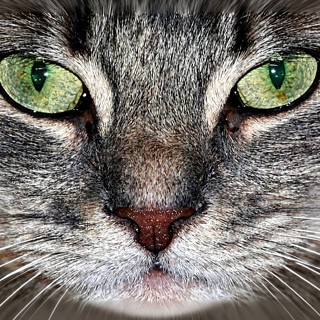}};
  \node[font=\tiny\bfseries, text=aegray!70!black, anchor=south, inner sep=1.5pt, align=center] at (endg.north)
    {fixed\\[-1pt]instruction\\[-1pt]{\mdseries baseline, no}\\[-1pt]{\mdseries human rating}};
  \draw[simlink, rounded corners=4pt] ($(sc.south)+(-0.12,0)$) |- (1.4,-3.95) -| (endb.south);
  \draw[-{Latex[length=1.5mm]}, draw=aegray, line width=0.6pt, rounded corners=4pt] ($(sc.south)+(0.12,0)$) |- (4.24,-3.95) -| (endg.south);

  \draw[black!20, line width=0.4pt] (0.2,-4.45) -- (5.44,-4.45);
  \finding{(0.25,-4.57)}{no}{No detected gain over a fixed instruction}
    {Under the loop's own scorer; no person judged the edits.}{5.1cm}
  \end{scope}
  \end{tikzpicture}
  \caption{Overview of the study on one EVA photograph. (a) A persona card is written from one rater's record. This rater gives the photograph 1 out of 10 while every persona gives 8, and profile regressions do not predict how held-out raters depart from the crowd. (b) The audience answers once for every real rating, and its scores bunch where real ratings spread out (schematic). A cross marks no detected advantage for the personas, and a half circle a partial one. (c) As an editing critic, the audience gets a kept edit with flat cartoon eyes, later replaced, and the right image is the fixed-instruction baseline's edit. No person evaluated any edit, so neither image stands for a human judgment of quality.}
  \label{fig:overview}
\end{figure*}

Language models that are told to answer as a particular kind of person are increasingly used in place of human participants, to fill in surveys \cite{argyle2023}, to voice the opinions of demographic groups \cite{santurkar2023whose} and to simulate individuals whose answers can be studied before anything reaches real people \cite{park2026simulations}. Vision-language models (VLMs) extend this idea to visual taste, since a model asked to rate an image as a viewer of a given age, nationality or interest in art could stand in for that viewer. Many such calls could then form a synthetic audience whose verdict predicts how a group of people will react to an image and whose complaints tell an editor what to change. Such an audience would be cheap and fast and would never tire, which makes it attractive wherever human judgment is slow or costly to collect.

When such an audience agrees with real viewers, however, the agreement does not show that the personas are responsible, because several cheaper mechanisms can produce it. Simulated answers vary less than real ones \cite{bisbee2024synthetic}, so averaging many of them and correcting how the model uses the rating scale can match the average viewer with little help from the personas. A persona can follow the specific raters it was written from rather than the group they belong to, and it can turn a group into a caricature \cite{cheng2023compost} or flatten and misportray it \cite{wang2025misportray}. Comments that sound like the persona may only repeat the words of the prompt, and when the audience steers a generator, any gain can come from the loop's choice among candidates rather than from what the audience says. Aesthetic taste sharpens these questions, because viewers disagree about images for reasons that go beyond noise \cite{uma2021, plank2022}, and personal taste is usually learned from each viewer's own ratings rather than from a description of them \cite{ren2017personalized, zhu2022meta, pamela2026}. Telling these explanations apart is an attribution problem, in which every kind of agreement needs the control that could explain it away.

We study this problem for visual taste with a synthetic audience in which every persona is a short card written from the recorded profile of one real rater. We draw the profiles from three datasets that record who rated each photograph or painting \cite{yang2022para, kang2020eva, maerten2025lapis}, add a crowd's votes on pairs of generated images \cite{rapidata2024}, and ask one frozen VLM, Qwen2-VL-7B-Instruct with greedy decoding of ratings, to answer once for every human judgment. We study a synthetic visual audience, in which independent, profile-conditioned viewers respond to shared images and their judgments are aggregated by the observed composition of the human panel. This is a setting for population construction and validation, not a model of interpersonal influence, and any interacting simulation built from such personas would start from this population. That demographic prompts can fail is known \cite{aesbiasbench2025, hu2024persona}, and what we add is an integrated audit in which each kind of agreement is paired with a comparator, namely a rater's own ratings of other images for stable individual offsets, a calibrated persona-free panel of the same size for image means, disjoint rater halves and centered gaps for group differences, masked comments for persona language, and a fixed instruction under the same selector for editing. In this configuration a cheaper comparator matches or beats most of the agreement, and we report these negative results as such, together with the one positive result on paintings. Our contributions are as follows.
\begin{itemize}
\item A comparator-per-claim audit for synthetic populations. Five kinds of agreement (single ratings, image means, group gaps, persona language and downstream editing) are each paired with the cheaper explanation that must be ruled out, and with the validity it can and cannot support (Table~\ref{tab:protocol}).

\item An application to one frozen, greedily decoded VLM configuration. After calibration a persona-free prompt matches photo means better, and the persona prompt does better on paintings under a confounded comparison. Group agreement is largely an image-independent offset, and it reverses on EVA.

\item Evidence that persona language and optimization gains do not establish fidelity. Attributes stay decodable from comments where ratings invert the real gap, and the audience critic shows no detected gain over a fixed instruction under the selector's own scorer.
\end{itemize}

\begin{table*}[t]
  \centering
  \footnotesize
  \setlength{\tabcolsep}{7.3pt}%
  \begin{tabular}{llccccccc}
    \toprule
    & & & & \multicolumn{2}{c}{Reliability $\uparrow$} & \multicolumn{3}{c}{Rater deviation explained $\uparrow$} \\
    \cmidrule(lr){5-6}\cmidrule(lr){7-9}
    Dataset & Content & Images & \hd{Raters\\per image} & \hd{single\\rating} & \hd{group\\average} & \hd{full profile,\\new raters} & \hd{best attribute,\\new raters} & \hd{own ratings,\\new images} \\
    \midrule
    PARA & photographs & 2{,}000 & 26 & 0.470 & 0.958 & $-$1.5\% & age 0.4\% & \textbf{18\%} \\
    EVA & photographs from AVA & 4{,}070 & 34 & 0.223 & 0.906 & $-$3.4\% & none & \textbf{31\%} \\
    LAPIS & paintings & 4{,}000 & 23 & 0.188 & 0.839 & $-$0.9\% & interest in art 1.8\% & \textbf{19\%} \\
    \bottomrule
  \end{tabular}
  \caption{The rating datasets, and what predicts how a rater departs from the crowd. Reliability is the one-way intraclass correlation of a single rating and of an image's mean rating. The last three columns give the out-of-sample share of residual variance (a rating minus its image mean) explained beyond a constant, by the full profile and by its best attribute on held-out raters, and by the rater's mean residual on other images (Section~\ref{sec:ceiling}). A negative share is worse than a constant.}
  \label{tab:datasets}
\end{table*}

\section{Related Work}

\textbf{Simulating People with Language Models.} Language models conditioned on a person's demographics can reproduce the distribution of survey answers of many groups \cite{argyle2023}, which has made persona-conditioned models a popular proxy for human samples. Later work found that simulated answers vary less than real ones \cite{bisbee2024synthetic}, that the opinions of several groups are misrepresented \cite{santurkar2023whose}, that persona prompts produce caricatures \cite{cheng2023compost}, and that models standing in for identity groups both misportray and flatten them \cite{wang2025misportray}. Agents built from interviews reproduce individuals' answers more accurately than agents built from demographic descriptions \cite{park2026simulations}, which suggests that a short profile describes a person only weakly. Persona variables also explain little of the variance in subjective text annotations \cite{hu2024persona}, and group-level sociodemographics do not capture individual annotation behavior \cite{orlikowski2023ecological}. We ask the same questions of visual taste, where repeated ratings from the same people separate a rater's stable offset from what a profile regression recovers. Among the fidelity criteria of \citet{argyle2023}, our comment test checks backward continuity and our group-gap test pattern correspondence, and EVA shows that the first can hold while the second fails.

\textbf{Personas and Visual Aesthetics.} Aesthetic scorers predict one consensus score per image, or the spread of a crowd's scores without saying who gives which \cite{talebi2018nima, schuhmann2022, wu2024qalign}, and general VLMs remain weaker than people at detailed aesthetic judgment \cite{huang2024aesbench}. Models of individual taste learn from each user's own ratings \cite{ren2017personalized, zhu2022meta, pamela2026}, which a profile does not contain. P-MLLM trains visual-fusion components to read profiles and evaluates user-specific correlations on PARA and LAPIS \cite{wang2026pmllm}, whereas we audit a frozen model that receives only a card. AesBiasBench shows that identity information in the prompt shifts how multimodal models judge images \cite{aesbiasbench2025}, whereas we ask whether such shifts match the gaps between real raters' groups, and the widely used LAION aesthetic predictor favors particular kinds of images and artists \cite{taylor2026gaze}. We add controls that separate calibration, fixed group offsets, image-specific group gaps and comment language.

\textbf{Critique and Selection in Generation Loops.} Generative outputs are improved with learned rewards \cite{xu2023imagereward, black2024ddpo, wallace2024dpo} or with language feedback \cite{madaan2023selfrefine, wu2024sld, wang2024genartist, yang2024idea2img}, for images often through an instruction-based editor \cite{brooks2023instructpix2pix, flux2024}. ZIPP has a language model playing a persona rewrite a generator's prompt for that viewer \cite{si2026zipp}, whereas we test whether a persona's critique adds anything over a fixed instruction under the same selector. When the same model proposes and judges, the reported gain can reflect its own bias \cite{xu2024selfbias}, which a jury of models from different families reduces \cite{verga2024juries}. A loop that keeps the best of several candidates also mixes the contributions of its critic and its selector, so we compare critics under the same selector, with paired uncertainty.

\section{The Audience and the Editing Loop}
\label{sec:method}

\begin{table*}[t]
  \centering
  \footnotesize
  \begin{tabularx}{\textwidth}{>{\raggedright\arraybackslash}p{0.085\textwidth}>{\raggedright\arraybackslash}p{0.19\textwidth}>{\raggedright\arraybackslash}X>{\raggedright\arraybackslash}p{0.155\textwidth}>{\raggedright\arraybackslash}p{0.155\textwidth}}
    \toprule
    Agreement & Comparator & Observed result & Validity it can support & Not identified \\
    \midrule
    \textbf{Single ratings} & Calibrated persona-free prompt; profile-only regressions on human data & Persona and persona-free errors nearly equal on photos (0.126 against 0.122; 0.156 against 0.155), slightly lower with personas on paintings (0.212 against 0.216). Profiles predict held-out raters' constant offsets worse than a constant; a rater's own history explains 18 to 31\% & Individual-level criterion validity: not shown & Profile-by-image interactions; the card's effect apart from the prompt framing \\
    \rowcolor{aerowalt}
    \textbf{Image means} & Persona-free prompt with the same held-out calibration; priors fitted on training halves & Persona-free prompt more accurate on PARA (by 0.0061) and EVA (by 0.0026); persona prompt more accurate on LAPIS (by 0.0065) & Held-out-image accuracy after calibration on human labels & The cards' contribution (the prompts differ); use without human labels \\
    \textbf{Group gaps} & Disjoint rater halves within each image; centering; lookup of measured gaps & Interest in art 0.229 uncentered, 0.032 centered; PARA photography experience 0.034; EVA gaps negative; the lookup matches or beats the audience on 7 of 8 tests & Descriptive agreement with observed groups, mostly as offsets & Causal demographic effects; generalization to raters held out globally \\
    \rowcolor{aerowalt}
    \textbf{Comments} & Card words masked; folds split by rater; persona-free classifier & AUC 0.58 to 0.70 after masking; persona-free classifier at chance & Attribute information in the generated text (a manipulation check) & Human-like language; faithful reasons or ratings \\
    \textbf{Editing} & Fixed instruction, persona-free critic and generic-instruction cycle, all under the same selector and gate & $+$0.006 \ci{$-0.020$}{0.034} over the fixed instruction; the generic cycle reaches 0.187; 110 extra VLM calls per image & Performance under the selector's own score & Benefit to people or a target group; robustness across seeds \\
    \bottomrule
  \end{tabularx}
  \caption{The audit as a checklist. Each row pairs a kind of agreement with real viewers with the comparators that could explain it, the observed result in this configuration, the validity the comparison can support, and what it leaves unidentified.}
  \label{tab:protocol}
\end{table*}

Figure~\ref{fig:overview} gives an overview of the study, from the synthetic audience and its tests to its use as an editing critic.

\textbf{Data.} We use three datasets in which each rating is tied to the person who gave it and to what they reported about themselves (Table~\ref{tab:datasets}). PARA \cite{yang2022para} contributes 2{,}000 photographs with 51.7k ratings on a 1 to 5 scale, EVA \cite{kang2020eva} 4{,}070 photographs drawn from AVA \cite{murray2012ava} with 137k ratings from 0 to 10, and LAPIS \cite{maerten2025lapis} 4{,}000 paintings with 90k ratings from 0 to 100. We also use 898 pairs of generated images from Rapidata \cite{rapidata2024} with 23{,}445 preference votes, for which only each voter's country and language are known. We study the overall aesthetic score and rescale every score linearly to $[0,1]$ using the ends of its scale.

\textbf{Persona Cards.} A persona card is a short text filled in from one real rater's record with a fixed template, and it contains only what the dataset records, namely age, gender, education, art and photography experience and Big-Five personality on PARA; age, gender, region, photographic level and eyesight on EVA; and nationality, age, gender, education, interest in art and eyesight on LAPIS. A PARA card, for example, begins \emph{``a female aged 22-25, education level: university; art experience: competent; photography experience: beginner''} and continues with five personality scores. Every persona is the card of a real rater, and our claims are limited to what these cards contain (templates in Appendix~\appref{sup:prompts}{H}).

\textbf{The Model.} We query \texttt{Qwen2-VL-7B-Instruct} \cite{wang2024qwen2vl} in bfloat16 precision, and no generative-model weights are updated, although calibration maps and diagnostic predictors are fitted. Each call receives one image, one persona card and the rating question of its dataset, which states the scale and asks for a score and a one-sentence comment in a fixed JSON format, and the audience makes one call for every human rating. The persona-free panel asks the same question without a card, also once per rating, so both panels have the same size, and every comparison between them uses the ratings both cover. Ratings are decoded greedily. A requested temperature of 0.7 was overridden by the model's shipped generation settings (top-$k$ of 1, Appendix~\appref{sup:repro}{K}), so both persona runs are in effect greedy. We report the second, whose settings match the persona-free panel's, and give the first in Appendix~\appref{sup:robust}{D}. The two runs differ on 12 to 21\% of matched ratings, which is consistent with numerical nondeterminism in batched bfloat16 inference, although we did not isolate the source. Under greedy decoding, a card that shifts probability without changing the most likely score leaves no trace, so our tests bound the card's effect on the modal answer only. A greedy persona-free panel gives nearly one answer per image, so it amounts to a single judgment and cannot separate groups by construction. The Rapidata votes are sampled at a temperature of 0.7, with the order of the two images drawn at random for each vote. The persona and persona-free prompts differ in role framing as well as in the card. The persona-free prompt asks for a typical viewer's rating, which may favor it on image means, and the persona prompt asks for in-character comments and lets the attributes shape ``how critical or generous'' the viewer is. The LAPIS prompts also differ from the photographic ones in taste wording, score resolution and output fields (Appendix~\appref{sup:prompts}{H}), and we do not measure the contribution of any of these differences. Rater halves are drawn within each image and group cell, and target-half raters' labels on other images can enter the calibration fitted on those images.

\textbf{Calibration.} The model rates almost every image within a narrow band of its scale, mostly above the human average, and places 34 to 56\% of its answers on its most common value, so we calibrate each rating with isotonic regression \cite{niculescu2005}, a monotone map from the raw score to the human rating of the same rater and image, pooled over raters. We fit the map on the human ratings of one random half of the images and apply it to the other half, and the reverse, calibrate the persona-free panel on the same halves, and take the mean of a group's calibrated ratings as its prediction. Every calibrated result is thus a held-out-image prediction after calibration on human ratings.

\textbf{Guiding Image Editing.} As a downstream use, the audience's complaints about the current best image become one instruction for an image editor, FLUX.1-Kontext-dev \cite{flux2024}. The editing loop uses 100 photographs (50 from PARA, 50 from EVA) and a fixed panel of ten PARA-derived cards for every image, whose scores are logged but never select an edit. The editor applies the instruction to the original photograph with three seeds per round, and a candidate is eligible only if its DINOv2 similarity to the original is at least 0.78. The LAION aesthetic predictor, on a scale of about 1 to 10, keeps the best eligible candidate if it beats the best image so far, and the same predictor measures every gain, so editing results are internal to this loop. EVA images come from AVA, on which the LAION predictor was trained. The instruction is merged with the previous one in every round, so requests from unsuccessful rounds can persist. Over ten rounds, we compare the audience with three other critics, namely one call without a persona (the persona-free critic), a fixed instruction, and eight generic instructions cycled under the same selection (the generic-instruction cycle). Each condition ran once with a fixed seed schedule, and Appendix~\appref{sup:loop}{F} gives the loop in full.

\textbf{Statistics.} Intervals are 95\% bootstrap intervals over images, or over raters for the comment classifier, paired whenever two conditions share images. The Holm correction covers the eight uncentered other-half group tests, and the centered, own-half and aspect results are exploratory. All intervals are conditional on the observed raters, the model configuration and, for editing, the seed schedule.

\section{What Explains the Audience's Agreement with Real Viewers}

Table~\ref{tab:protocol} lists the kinds of agreement with real viewers that we test, the comparator for each and what each comparison can support. This section takes the first four in turn, from single raters to groups and comments, and Section~\ref{sec:c4} turns to editing, where the audience acts as a critic.

\subsection{Profile Regressions Do Not Predict How a Viewer Departs from the Crowd}
\label{sec:ceiling}

Before testing the model, we ask how much a profile could reveal about one person's ratings at all. The intraclass correlation of a single rating is the expected agreement between two people rating the same images, and that of a group average is the expected agreement between the averages of two separate groups of raters \cite{shrout1979}. We then treat each rating as the image's mean rating plus a residual, and predict the residuals in three ways, by a linear regression on the one-hot encoded recorded profile fitted with five folds split by rater and scored on raters it never saw, by the same regression on the single best attribute, and by the rater's own mean residual on the other half of the images. We score each prediction by the share of the residual variance that it explains beyond a constant fitted on the same training folds. Since each prediction is one number per rater, it captures only the part of a viewer's departure from the crowd that holds across images, which mixes preference with how they use the scale, and it does not test profile-by-image interactions.

Table~\ref{tab:datasets} shows that a group average is far more reliable than a single rating, at 0.84 to 0.96 against 0.19 to 0.47, so most of the variance in one rating is not shared with other viewers. A stable component of each rater's departure from the crowd is predictable from their other ratings, since a rater's mean residual on other images explains 18 to 31\% of the residual variance. The tested profile regressions capture none of this for held-out raters, since the full profile predicts worse than a constant, by 0.9 to 3.4\% of the variance, and ridge regression gives the same result. Single attributes do little better, since the strongest, interest in art on LAPIS, explains 1.8\%, age on PARA 0.4\% and no attribute on EVA anything. The audience's single ratings are also no more accurate than the persona-free panel's on photographs (Appendix~\appref{sup:calib}{B}), so the rest of this section asks what the audience reproduces for groups of viewers rather than for single raters.

\subsection{After Calibration, a Persona-Free Judge Matches Photo Means Better}
\label{sec:generated}

\begin{table}[t]
  \centering
  \footnotesize
  \setlength{\tabcolsep}{5.4pt}%
  \begin{tabular}{lccccc}
    \toprule
    & \multicolumn{2}{c}{Audience $\downarrow$} & & \multicolumn{2}{c}{Prior $\downarrow$} \\
    \cmidrule(lr){2-3}\cmidrule(lr){5-6}
    Dataset & raw & calibrated & \hd{persona-free\\calibrated $\downarrow$} & mean & median \\
    \midrule
    PARA & 0.182 & 0.062 & \textbf{0.056} & 0.103 & 0.112 \\
    EVA & 0.110 & 0.063 & \textbf{0.061} & 0.081 & 0.082 \\
    LAPIS & 0.161 & \textbf{0.073} & 0.079 & 0.105 & 0.108 \\
    \bottomrule
  \end{tabular}
  \caption{After calibration on human ratings, a persona-free panel matches photo means better. Mean absolute error of each image's mean rating on held-out image halves, on the $[0,1]$ scale, with the lowest error in bold. Priors are fitted on the training half of the images, and the values match the overall rows of Table~\appref{tab:aspects}{9}.}
  \label{tab:calibration}
\end{table}

We first ask whether the audience predicts the average rating of an image, the quantity that a single aesthetic score targets. We compare the mean absolute error of the audience's calibrated group mean with that of its raw mean, of priors that always predict the mean or median rating, and of the persona-free panel calibrated in the same way. On the generated pairs of Rapidata, the model casts one vote for every human vote under that voter's card, and we compare the majority of these votes with the crowd's majority, against a popularity prior that always picks the image from the generator the crowd prefers overall.

Table~\ref{tab:calibration} shows that raw image-mean predictions are less accurate than constant priors on every dataset. The audience is also flatter than the panel it simulates. The model places 34 to 56\% of its answers on its most common value, against 5 to 30\% for people, and on PARA and EVA all personas give the same score on 42 to 53\% of images across our two runs, so adding personas often cannot move the mean at all. This is the flattening reported for persona-conditioned language models \cite{bisbee2024synthetic, wang2025misportray}, and greedy decoding contributes to it. Calibration cuts the error by 42 to 66\% and beats both priors on every dataset, but the persona-free panel, calibrated in the same way, is more accurate still on PARA, where the persona panel's error is higher by 0.0061~\ci{0.0044}{0.0079}, and on EVA, where it is higher by 0.0026~\ci{0.0017}{0.0034}. Only on the LAPIS paintings is the persona prompt more accurate, lowering the error by 0.0065~\ci{0.0053}{0.0076}, less than a tenth of what calibration removes there. Since the greedy persona-free panel amounts to one judgment, a single calibrated judgment without any profile matches the average viewer of a photograph better than a panel of 26 to 34 personas. The LAPIS exception comes with a different prompt, score resolution and output schema, so we cannot attribute it to the cards alone. The 16 rated aspects show the same pattern (Appendix~\appref{sup:robust}{D}).

On the generated pairs, the audience's majority agrees with the crowd's majority on 65.9\%~\ci{62.6}{68.9} of pairs, against 60.0\%~\ci{56.9}{63.1} for a popularity prior that never looks at the images. Under the split-half comparison, in which the audience casts only the votes of half of the voters and both it and the prior are scored against the other half, the observed advantage falls to 0.4 percentage points (61.9 against 61.5\%, no paired interval computed), while two crowd halves agree on 67.3\%. We have no persona-free panel on these pairs, so even this cannot be attributed to the personas.

\subsection{Art-Interest Agreement Is Strongest and Largely Image-Independent}
\label{sec:audience}

\begin{figure}[t]
  \centering
  \includegraphics[width=\linewidth]{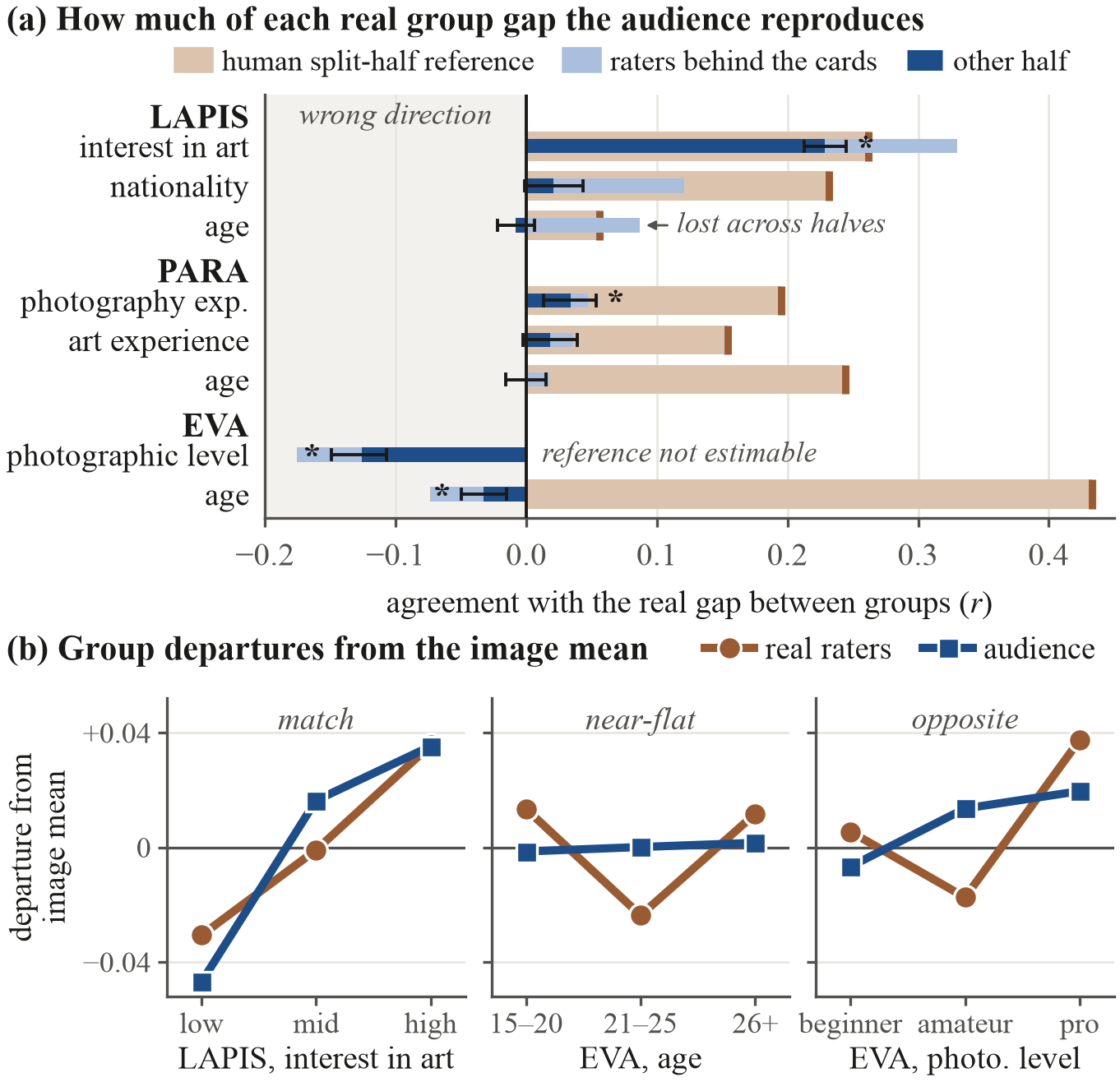}
  \caption{Group-gap agreement. (a) Uncentered correlation of predicted with real gaps between viewer groups, against all raters behind the cards (light bars) and against the other half of the raters (dark bars, with 95\% intervals over images, $\ast$ when the value survives the Holm correction). Brown bands run to the human split-half reference. Light bars are scored against all raters, a larger and less noisy target, so they can exceed the half-sample reference. Centered values are in Table~\appref{tab:validity}{7}. (b) Each group's departure from the image mean, real and predicted. On EVA age the predictions are near-flat, and their gaps take the opposite sign to the real ones in (a).}
  \label{fig:validity}
\end{figure}

Groups of viewers rate differently, since on paintings viewers with high interest in art rate higher, and on EVA viewers aged 21 to 25 are the harshest (Figure~\ref{fig:validity}b). Across disjoint rater halves, the audience's predicted gaps for interest in art agree with the real ones at 0.229, but only at 0.032 once each group pair's average gap is removed. Photography experience on PARA agrees weakly (0.034), and both EVA attributes run negative, meaning the audience reverses the real gaps.

For each of eight attributes across the three datasets, we split the raters of each image into groups, cutting numeric attributes into thirds and keeping groups of at least four raters on that image, and take the gap between the mean ratings of every pair of groups present on it, in the real and in the calibrated predicted ratings. Over all such cells, unweighted, we compute the uncentered correlation $\sum pq/\sqrt{\sum p^2 \sum q^2}$ of predicted gaps $p$ with real gaps $q$, so that predicting the direction of a gap that holds on every image counts as agreement, and its centered version, which first removes each group pair's average gap over images from $p$ and $q$ separately and so scores only how gaps vary between images. Scored against the raters whose cards it received, the audience can succeed by following those specific raters, so we also split the raters of every cell into two random halves, 20 times, and score the prediction for half A against the real ratings of half B. A human split-half reference, $\sqrt{\sum q_A q_B/\sum q_B^2}$, is a reference for the best attainable correlation against half B if the two halves' sampling errors are independent.

Figure~\ref{fig:validity}a shows that most of the agreement weakens across halves (values in Appendix~\appref{sup:robust}{D}). Against the raters it was given, the audience is positive on all six tests on LAPIS and PARA, but across halves LAPIS nationality falls from 0.121 to 0.021, LAPIS age from 0.087 to $-0.008$ and PARA age from 0.017 to 0.000. Part of such a drop could come from the smaller target, so we also score the prediction for half A against half A itself. On these three tests the own half scores higher than the other half by 0.071, 0.105 and 0.024, with intervals that exclude zero, so agreement there depends on scoring against the same within-image raters, which does not by itself show recovery of individual taste. Interest in art on LAPIS is the strongest test, at 0.229~\ci{0.213}{0.245} across halves and 0.227 against the own half, so its drop from 0.330 reflects the smaller target rather than rater-specific tracking, and its split-half reference is 0.262. Most of this agreement is a fixed offset, since viewers with high interest rate paintings higher across the board (Figure~\ref{fig:validity}b), and the centered correlation is only 0.032~\ci{0.017}{0.046} across halves. The real gaps are themselves mostly an offset, since the two human halves' centered gaps do not covary ($-0.003$), so no centered reference can be estimated. PARA photography experience also survives the Holm correction, at 0.034, and no centered value exceeds 0.032.

Elsewhere the audience misses real group gaps or reverses them. On PARA, age has a split-half reference of 0.244, yet the audience reproduces none of it across halves. On EVA, both tests are significantly negative across halves, at $-0.126$ for photographic level and $-0.033$ for age. The real pattern there is U-shaped (Figure~\ref{fig:validity}b), with viewers aged 21 to 25 and amateur photographers the harshest, whereas the model's predictions rise with age and experience, and its range across ages is about a twelfth of the real one. The prompt's invitation to let attributes set generosity is one plausible source of such a monotone rule, which we do not isolate. A lookup of group gaps measured on other raters and images, scored on the same cells as the audience, is higher by 0.08 or more on four tests, within 0.012 on three and lower only on LAPIS age, where the audience is scored on the raters it was given. The persona-free panel stays near zero on every test, as expected from a panel whose answers barely vary, and the two persona runs differ by at most 0.01 across halves (Appendix~\appref{sup:robust}{D}).

\subsection{Attributes Are Decodable from Comments, Not Validated by Them}
\label{sec:persona}

Finally, we ask whether the comments the audience writes carry the persona. A logistic regression on the words of each comment predicts one attribute of the rater it was written for, namely art experience on PARA, photographic level on EVA and interest in art on LAPIS, with folds split by rater so that no rater appears in both training and testing. To test whether the comments only repeat the card, we mask every word that appears in any card of the dataset, 42 to 125 words including every age, nationality and level, and fit the classifier again.

The classifier recovers the attribute above chance on every dataset, with an area under the ROC curve (AUC) of 0.58~\ci{0.58}{0.59} on PARA, 0.70~\ci{0.69}{0.72} on EVA and 0.70~\ci{0.68}{0.71} on LAPIS, against 0.50 for the comments of the persona-free panel, and masking every card word moves these values by at most 0.0003 (Appendix~\appref{sup:probe}{C}). Card words occur in 20\% of PARA comments, 5\% of LAPIS comments and 0.3\% of EVA comments, so on EVA the masking test has little to remove, and paraphrases of the card are not ruled out. The prompted attribute is therefore decodable from the generated comments, which we cannot compare with human comments, since the datasets record none. Set against the previous sections, this warns against judging a persona by its words, since on EVA the comments reveal a viewer's photographic level with an AUC of 0.70 while the ratings reverse the real difference between levels.

\section{Guiding Editing: Critique or Selection?}
\label{sec:c4}

\begin{figure}[t]
  \centering
  \includegraphics[width=\linewidth]{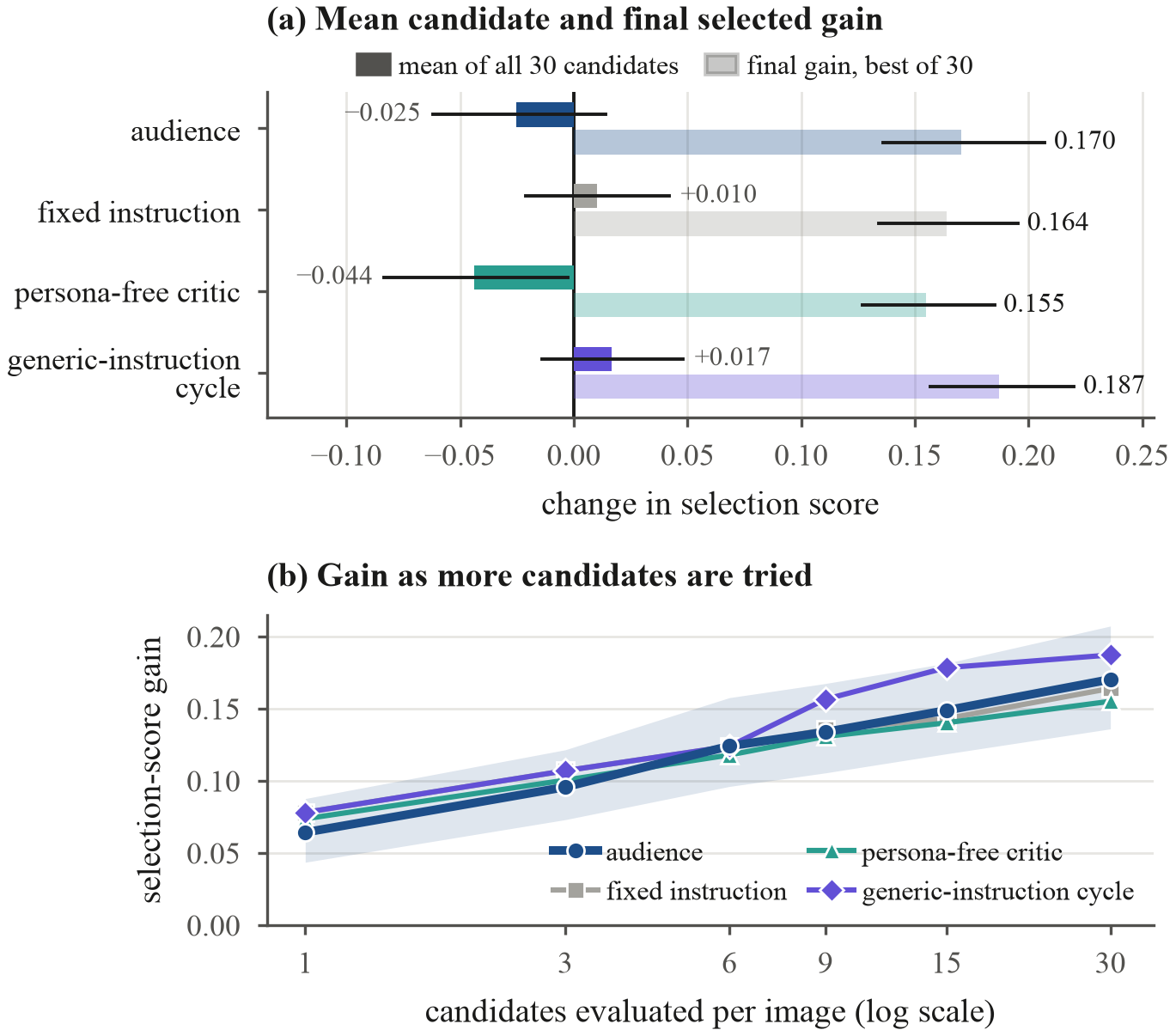}
  \caption{Editing gains under the loop's own scorer. (a) The mean change of all 30 logged candidates from the source (dark), a mean along the adaptive trajectory rather than a run without selection, and the final selected gain (light), with 95\% intervals over images. (b) Mean selected gain after the first $n$ logged candidates, with a 95\% band for the audience.}
  \label{fig:selection}
\end{figure}

We now use the audience for what a synthetic audience is often wanted for, as a critic that tells an image editor what to change. Because the loop keeps an edit only when the scorer prefers it, a gain can come from the critique or from the choice among the 30 candidates of each image, so we compare critics under the same selector and gate, pair them by image, and describe the logged candidates (Appendix~\appref{sup:loop}{F}).

At 30 candidates the audience's selected gain, 0.170, exceeds the fixed instruction's by 0.006~\ci{$-0.020$}{0.034}, so we detect no incremental benefit, and the interval does not establish equivalence. The persona-free critic gains 0.155, a difference of 0.015~\ci{$-0.010$}{0.042}, and at every evaluated budget the interval of the audience's paired difference from each other critic includes zero (Figure~\ref{fig:selection}b). The generic-instruction cycle, which never sees the image, reaches 0.187, which is consistent with selection rather than critique driving the gains. Mean candidate changes along the logged trajectories, whose later instructions depend on selection, lie within 0.05 of zero for every critic (Figure~\ref{fig:selection}a, Appendix~\appref{sup:loop}{F}). The similarity gate passes 85\% of the audience's candidates against all of the fixed instruction's, and 7 of the audience's images have no eligible candidate in any round and keep the source.

The audience also changes the kind of edit requested. A keyword tagger labels the instruction behind 44\% of its kept edits (75 of 171) as adding, removing or replacing an object, against 25\% for the persona-free critic and 2\% for the generic-instruction cycle, at a cost of 110 extra VLM calls per image against 20 and none, on top of the 30 edits and scorer passes that every critic shares. Such edits can undo what made a photograph distinctive, as with the cat's flat cartoon eyes kept at round 2 (Figure~\ref{fig:overview}). These results are limited to the implemented scorer, gate and instruction history, and no person judged whether the edits serve any target group.

\section{Discussion}
\label{sec:discussion}

As Table~\ref{tab:protocol} sums up, calibrated aggregate accuracy, group offsets, attribute-decodable language and internal editing gains each support a different kind of validity. Including the painting result, these comparisons do not isolate the cards' contribution or establish individual fidelity. The cross-half test detects one group offset, so it is not blind by construction, but no comparable positive case exists for the centered test. We draw three lessons for anyone who builds a synthetic audience. First, group agreement should be reported across disjoint rater halves and after centering, since agreement with the raters behind the cards partly reflects those raters and an uncentered correlation credits a fixed offset. Second, every persona panel should be compared with a calibrated persona-free panel of the same size, since calibration on human ratings recovers most of what a panel gets right about image means. Third, a loop that uses the audience to steer generation should be compared with a fixed instruction under the same selector and judged by an outside evaluator.

\section{Limitations}

Our empirical conclusions are limited to this frozen Qwen2-VL-7B-Instruct configuration and its prompts, and the comparators, not the numbers, are meant to transfer. A panel whose personas often agree exactly is itself a limit of audiences built this way, to which greedy decoding contributes. The audiences represent the raters retained in each dataset, not national or demographic populations. The personas are only as rich as the records, and richer descriptions such as interviews may carry more of a person \cite{park2026simulations}. Our audience is a population of independent raters whose judgments are aggregated, so we do not test whether interaction among simulated viewers would change these conclusions. The prompts differ between conditions and datasets in more than the card, and rewording alone can change a model's answers \cite{rezaei2026fragility} and judgments \cite{subasic2026lottery}. Every calibrated result also uses human ratings. Pretraining exposure to these images or ratings would inflate agreement, so it cannot explain the null results, although it could contribute to the painting result. In editing, every gain is measured by the scorer that selects the edits, each critic ran once with a fixed seed schedule, and no person judged the edits. A persona audience can also misrepresent the groups it claims to stand for, as the reversed age pattern on EVA shows, so it should be tested against the population it is meant to represent before anyone relies on its verdicts.

\section{Conclusion}

For one frozen VLM configuration, every kind of agreement we tested had a cheaper explanation or no detected advantage, and the painting result cannot be assigned to the cards alone. As a checklist, a synthetic population needs five comparators, namely a calibrated persona-free prompt and profile-only regressions for single ratings, the same held-out calibration without cards for image means, disjoint and centered rater halves for group gaps, masked card words and a persona-free classifier for comments, and a fixed instruction under the same selector for downstream use. Agreement with human ratings is not sufficient evidence of persona fidelity, and claims that a synthetic audience simulates viewers, demographic groups or useful feedback should come with comparators like these, reported beside every agreement.

\bibliography{references}

\ifdefined\mainonly\else
\appendix
\section*{Appendix}

This appendix gives further detail behind the main paper. Appendix~\ref{sup:data} summarizes the four datasets, Appendix~\ref{sup:calib} explains why the audience's ratings need calibration and how well a calibration carries over between datasets, and Appendix~\ref{sup:probe} describes the classifier that reads a rater's attribute from the comments. Appendix~\ref{sup:robust} repeats the group results on the first persona run and gives the lookup comparison, and Appendix~\ref{sup:fairness} measures whether the model treats all kinds of viewers alike. Appendix~\ref{sup:loop} describes the editing loop in full, with a diagram and a worked example, Appendix~\ref{sup:c4qual} shows editing examples drawn at random, annotated examples where the audience did worst and selected examples where it did best, and Appendix~\ref{sup:prompts} gives every prompt verbatim. We close with future directions, ethics and reproducibility. Unless stated otherwise, the rating results in this appendix come from the first of the two greedy persona runs, so they differ from the main text, which reports the second, by up to 0.017 in raw error but by less than 0.001 after calibration, and every confidence interval comes from 1{,}000 bootstrap resamples of images or raters, as stated with each result.

\section{Datasets}
\label{sup:data}
Table~\ref{tab:data} summarizes the four datasets described in Section~\ref{sec:method}, with the number of images and ratings we use and what each dataset records about its raters. How much a dataset records does not predict where the audience works, since the strongest group-gap agreement across rater halves comes from a single attribute, interest in art on LAPIS (Section~\ref{sec:audience}). The reliabilities of Table~\ref{tab:datasets} are one-way random-effects intraclass correlations with raters nested in images, ICC(1) for a single rating and ICC($k$) for the mean of an image's ratings, with $k$ the average number of raters per image corrected for unequal counts. A residual is a rating minus the mean of all ratings of its image, and the best attribute is the single-attribute regression with the highest out-of-fold share, chosen on the same rater folds that score it, so its value is optimistic rather than a held-out estimate.

\begin{table*}[t]
  \centering
  \footnotesize
  \begin{tabularx}{\textwidth}{llcc>{\raggedright\arraybackslash}X}
    \toprule
    Dataset & Content & Images & Ratings & Rater information \\
    \midrule
    PARA \cite{yang2022para} & photographs & 2{,}000 & 51.7k & age, gender, education, art and photography experience, Big-Five personality \\
    EVA \cite{kang2020eva} & photographs from AVA & 4{,}070 & 137k & age, gender, region, photographic level, eyesight \\
    LAPIS \cite{maerten2025lapis} & paintings & 4{,}000 & 90k & nationality, age, gender, education, interest in art, eyesight \\
    Rapidata \cite{rapidata2024} & AI-generated pairs & 898 pairs & 23k votes & country and language of each voter, across 120 countries \\
    \bottomrule
  \end{tabularx}
  \caption{The four datasets, with the number of images and ratings we use and what each records about its raters.}
  \label{tab:data}
\end{table*}

\section{Calibrating the Audience's Ratings}
\label{sup:calib}
We examine why the VLM's raw ratings need the calibration of Section~\ref{sec:method}, what it fixes, and whether a calibration fit on one dataset can be reused on another.

\begin{figure}[!htb]
  \centering
  \includegraphics[width=\columnwidth]{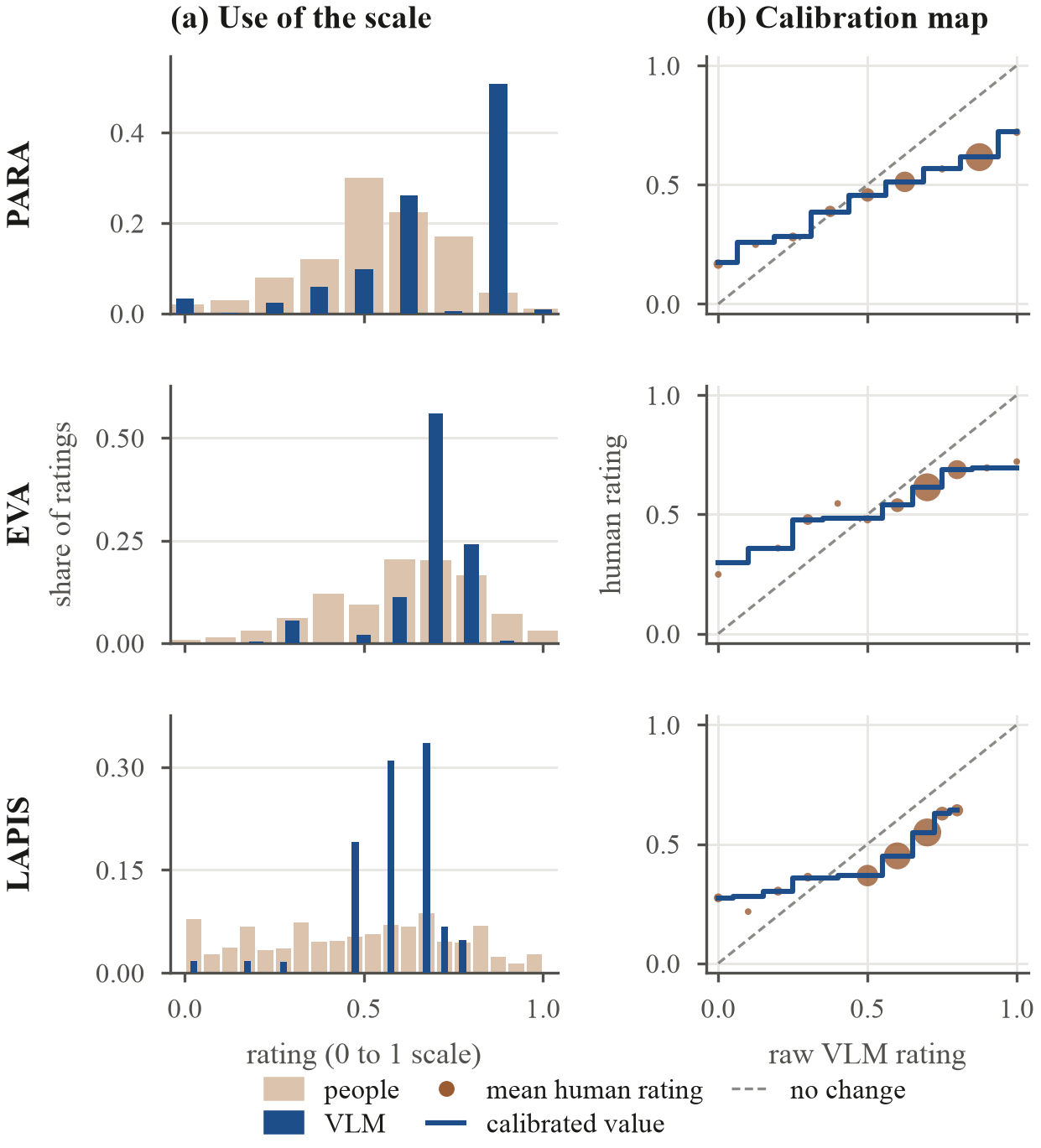}
  \caption{How the VLM uses the rating scale and what calibration does, with one row per dataset. (a) The share of ratings on each value of the scale for people (brown) and for the VLM (blue), with LAPIS binned in steps of 0.05. (b) For each raw VLM rating, the mean human rating of the same rater and image (brown, sized by the number of ratings) and the calibrated value used in the paper (blue), against the line of no change.}
  \label{fig:cal}
\end{figure}

Figure~\ref{fig:cal}a shows that the VLM places 34 to 56\% of its answers on the most common value against 5 to 30\% for people, and that on the wider scales of EVA and LAPIS it also uses fewer values, with an entropy of 1.7 and 2.3 bits against 3.0 and 4.2. Because this habit is a fixed trait of the model rather than noise, averaging over a panel cannot remove it, but a calibration that maps each raw rating onto the human scale can. Figure~\ref{fig:cal}b shows this map, which is flatter than the line of no change on every dataset, so calibration raises the VLM's lowest ratings and lowers its highest ones toward the human average.

In the second persona run, as in Section~\ref{sec:generated}, calibration cuts the error of the group average by 42 to 66\% but that of a single rating by only 8 to 39\%, leaving a single rating two to three times further from the truth, as expected from Section~\ref{sec:ceiling}, where most of the variance in one rating is not shared with other viewers. The personas do not narrow this gap, since on single ratings the calibrated audience is no more accurate than the persona-free panel, which gives nearly one answer per image, on the photographs (0.126 against 0.122 on PARA and 0.156 against 0.155 on EVA) and only slightly more accurate on the paintings (0.212 against 0.216).

\begin{table}[t]
  \centering
  \footnotesize
  \setlength{\tabcolsep}{10.0pt}%
  \begin{tabular}{lcccc}
    \toprule
    & & \multicolumn{3}{c}{Calibrated, fit on $\downarrow$} \\
    \cmidrule(lr){3-5}
    Evaluated on & Raw $\downarrow$ & PARA & EVA & LAPIS \\
    \midrule
    PARA  & 0.184 & \textbf{0.062} & 0.083 & 0.070 \\
    EVA   & 0.103 & 0.094 & \textbf{0.063} & 0.085 \\
    LAPIS & 0.145 & 0.086 & 0.100 & \textbf{0.072} \\
    \rowcolor{aerowmean}
    Mean  & 0.144 & 0.081 & 0.082 & \textbf{0.076} \\
    \bottomrule
  \end{tabular}
  \caption{Mean absolute error of the audience's average rating when the calibration is fit on one dataset (columns) and applied to another (rows), next to the raw error, in the first persona run, with the mean over the three datasets in the last row. A fit on another dataset always beats the raw error but does less well than a fit on the dataset itself, shown in bold, and a fit on LAPIS has the lowest mean error.}
  \label{tab:caltransfer}
\end{table}

Table~\ref{tab:caltransfer} shows that a calibration fit on another dataset always beats the raw error, for example 0.070 against 0.184 on PARA with a fit on LAPIS, though it never matches a fit on the dataset itself. A fit on one rated dataset can therefore be reused on another at some cost in accuracy, while we have not tested whether it carries over to generated images.

\section{Reading the Persona from the Comments}
\label{sup:probe}
Section~\ref{sec:persona} reports that the comments the VLM writes carry the persona it was given, even though its ratings often do not follow it. Here we describe the classifier behind that result and the data it is trained on, together with three checks of limited diagnostic purpose. Table~\ref{tab:probe} gives every number.

\begin{table*}[t]
  \centering
  \footnotesize
  \setlength{\tabcolsep}{8.2pt}%
  \begin{tabular}{llccccccc}
    \toprule
    & & \multicolumn{3}{c}{Persona AUC $\uparrow$} & \multicolumn{2}{c}{Persona-free AUC} & \multicolumn{2}{c}{Distinct comments} \\
    \cmidrule(lr){3-5}\cmidrule(lr){6-7}\cmidrule(lr){8-9}
    Dataset & Attribute & random & by rater & \hd{by rater,\\card words masked} & random & by rater & persona & persona-free \\
    \midrule
    PARA  & art experience     & 0.578 & 0.5785 & 0.5783 & 0.498 & 0.503 & 0.44 & 0.11 \\
    EVA   & photographic level & 0.701 & 0.7043 & 0.7045 & 0.511 & 0.513 & 0.24 & 0.08 \\
    LAPIS & interest in art    & 0.695 & 0.6953 & 0.6950 & 0.506 & 0.507 & 0.42 & 0.09 \\
    \rowcolor{aerowmean}
    Mean  &                    & 0.658 & 0.6594 & 0.6593 & 0.505 & 0.508 & 0.37 & 0.10 \\
    \bottomrule
  \end{tabular}
  \caption{Recovering a rater's attribute from the VLM's comment alone, with and without a persona, under random folds, folds split by rater, and with card words masked, in the first persona run, with the number of distinct comments per rater of an image. Chance is an AUC of 0.5, and art experience has four classes and the other attributes three.}
  \label{tab:probe}
\end{table*}

\subsubsection{Classifier and Data.} For each dataset, we pair every comment the VLM wrote while playing a rater with one attribute of that rater, namely art experience on PARA, photographic level on EVA and interest in art on LAPIS, and we discard the score so that the classifier sees only the text. Numeric attributes are split into three groups of equal size, categorical ones keep their own levels, and any level with fewer than 500 comments is dropped. We draw 30{,}000 comments at random from each dataset (seed 0), out of 51.7k on PARA, 136.9k on EVA and 90.3k on LAPIS, which come from 436, 998 and 530 raters. Each comment is turned into TF-IDF weights over the 2{,}000 most frequent words, ignoring common English words and words that appear in fewer than five comments, and a multinomial logistic regression from scikit-learn with the default L2 penalty ($C{=}1$, at most 300 iterations) predicts the attribute from these weights. We score it with three-fold cross-validation and report the one-versus-rest ROC-AUC averaged over classes, where 0.5 is chance. We use AUC rather than accuracy because the classes are uneven, most of all on EVA, where always guessing the largest class would already be right two times in three while its AUC would stay at 0.5.

\subsubsection{Attribute Information Accessible to a Lexical Classifier.} We keep the classifier deliberately simple, since it sees only which words appear in a comment, knows nothing about the image and draws a straight boundary between classes. Whatever it finds is therefore carried by plain word choice rather than by a high-capacity model fitting spurious features, so its scores measure the attribute information that word choice makes accessible, not everything the comments may carry in their phrasing, tone or structure.

\subsubsection{Persona-Free Classifier Control.} Raters with different attributes may have rated somewhat different images, and since the comments describe those images, a classifier could in principle guess the attribute from the image content alone. The persona-free panel tests this directly, because it sees the same images for the same raters and goes through the same pipeline but receives no persona. Its classifier stays at chance on every dataset, between 0.50 and 0.51, so the image content alone carries no usable trace of the attribute for this classifier.

\subsubsection{Folds Split by Rater.} Random folds place comments from the same rater in both the training and the test data, and because all of a rater's comments come from the same persona card, the classifier could learn the phrasing of particular cards instead of the attribute. We therefore repeat the evaluation with folds split by rater, so that no rater appears on both sides, and refit the TF-IDF weights inside each fold. The scores do not move, at 0.579, 0.704 and 0.695 against 0.578, 0.701 and 0.695 with random folds, and the persona-free panel stays at chance. The classifier has therefore learned how the model writes for a kind of viewer, and this carries over to raters it has never seen. Since the datasets contain no human comments, we cannot say whether real viewers of that kind write this way.

\subsubsection{Literal Card-Word Masking.} Comments could reveal the attribute simply by naming it, for example by stating an age or a level of interest in art. We therefore remove from every comment each word that appears in any card of its dataset, 42 words on PARA, 125 on EVA and 124 on LAPIS, including every age, nationality, region and level, and refit the classifier with folds split by rater. The AUC moves by at most 0.0003 on every dataset, in the first persona run as in the second, which gives 0.584, 0.701 and 0.696 before masking. The masked words are rare in the comments, since at least one occurs in 19.8\% of PARA comments, 5.1\% of LAPIS comments and 0.3\% of EVA comments in the second run, so on EVA the test removes almost nothing and its unchanged AUC says little. Masking also does not remove paraphrases of the card, such as a comment that says in other words that the viewer loves art, so part of the persona may still reach the comments in this way.

\subsubsection{Distinct Comments.} For every image rated by at least ten people, we divide the number of distinct comments by the number of raters and average this ratio over images. With personas, the VLM writes three to five times as many distinct comments per image as without them. Since every panel decodes greedily, the persona-free comments on one image differ only through numerical noise, so this ratio shows that different cards produce different text rather than that personas add variety which a sampled panel would lack.

\section{Robustness of the Results}
\label{sup:robust}
Table~\ref{tab:validity} gives the exact values behind Figure~\ref{fig:validity}.

\begin{table*}[t]
  \centering
  \footnotesize
  \setlength{\tabcolsep}{1.5pt}
  \begin{tabular*}{\textwidth}{@{\extracolsep{\fill}}llcccccccc@{}}
    \toprule
    & & \multicolumn{2}{c}{Raters behind the cards} & \multicolumn{2}{c}{Other half, 95\% CI} & \multicolumn{2}{c}{Split-half reference} & & \\
    \cmidrule(lr){3-4}\cmidrule(lr){5-6}\cmidrule(lr){7-8}
    Dataset & Attribute & \hd{all\\raters} & \hd{own\\half} & uncentered $\uparrow$ & centered $\uparrow$ & \hd{un-\\centered} & \hd{cen-\\tered} & Ratio & \hd{Persona-free,\\other half} \\
    \midrule
    LAPIS & art interest & $+$0.330 & $+$0.227 & $+$0.229$^\ast$ [0.213,\,0.245] & $+$0.032 [0.017,\,0.046] & 0.262 & n/a & 0.87 & $+$0.016 \\
    LAPIS & nationality & $+$0.121 & $+$0.092 & $+$0.021 [$-$0.002,\,0.044] & $-$0.019 [$-$0.040,\,0.002] & 0.231 & n/a & 0.09 & $-$0.001 \\
    LAPIS & age & $+$0.087 & $+$0.097 & $-$0.008 [$-$0.022,\,0.006] & $-$0.004 [$-$0.018,\,0.010] & 0.056 & n/a & n/a & $+$0.009 \\
    \midrule
    PARA & photo.\ exp. & $+$0.047 & $+$0.029 & $+$0.034$^\ast$ [0.013,\,0.053] & $+$0.019 [$-$0.002,\,0.039] & 0.195 & 0.126 & 0.17 & $-$0.008 \\
    PARA & art exp. & $+$0.036 & $+$0.025 & $+$0.018 [$-$0.002,\,0.039] & $+$0.014 [$-$0.006,\,0.034] & 0.154 & 0.114 & 0.12 & $-$0.011 \\
    PARA & age & $+$0.017 & $+$0.023 & 0.000 [$-$0.015,\,0.015] & $-$0.002 [$-$0.017,\,0.013] & 0.244 & 0.156 & n/a & $-$0.002 \\
    \midrule
    EVA & photo.\ level & $-$0.176 & $-$0.117 & $-$0.126$^\ast$ [$-$0.149,\,$-$0.107] & $-$0.014 [$-$0.036,\,0.007] & n/a & n/a & n/a & $+$0.017 \\
    EVA & age & $-$0.074 & $-$0.053 & $-$0.033$^\ast$ [$-$0.050,\,$-$0.015] & $-$0.027 [$-$0.043,\,$-$0.010] & 0.433 & 0.305 & n/a & $-$0.003 \\
    \bottomrule
  \end{tabular*}
  \caption{Group-gap agreement as the correlation of predicted and real gaps over all (image, group pair) cells, in the second persona run. All raters scores the prediction against every rater of a cell, and own half and other half score the prediction for half A against half A or half B, over 20 random splits (Section~\ref{sec:audience}). Centered removes each group pair's average gap. The split-half references assume independent half errors, the centered one applies the same formula after removing each group pair's average gap from both human halves, and the ratio divides the other-half value by the uncentered reference under that assumption. Intervals resample images, and $^\ast$ marks other-half values that survive the Holm correction over the eight uncentered tests. n/a marks a correlation that is not positive or a reference that cannot be estimated because the human halves' gaps do not covary, as for EVA photographic level ($r=-0.004$~\ci{$-0.032$}{0.026}) and every centered LAPIS reference ($r\le-0.002$). Over the 20 splits, the other-half value for interest in art ranges from 0.201 to 0.247.}
  \label{tab:validity}
\end{table*}

We check that the group results do not depend on which of the two persona runs we report, give the lookup comparison of Section~\ref{sec:audience}, compare the audience with the persona-free panel on every rated aspect, and report indirect checks related to memory of the images or their ratings.

\begin{table*}[t]
  \centering
  \footnotesize
  \begin{tabular*}{\textwidth}{@{\extracolsep{\fill}}llcccc@{}}
    \toprule
    & & \multicolumn{2}{c}{Other half $r$ $\uparrow$} & \multicolumn{2}{c}{Other raters and images $r$ $\uparrow$} \\
    \cmidrule(lr){3-4}\cmidrule(lr){5-6}
    Dataset & Attribute & first run & second run & lookup & audience \\
    \midrule
    LAPIS & interest in art & $+$0.219$^\ast$ & $+$0.229$^\ast$ & $+$0.245 & $+$0.233 \\
    LAPIS & nationality & $+$0.022 & $+$0.021 & $+$0.139 & $+$0.129 \\
    LAPIS & age & $-$0.003 & $-$0.008 & $+$0.062 & $+$0.105 \\
    \midrule
    PARA & photography experience & $+$0.032$^\ast$ & $+$0.034$^\ast$ & $+$0.139 & $+$0.041 \\
    PARA & art experience & $+$0.014 & $+$0.018 & $+$0.028 & $+$0.029 \\
    PARA & age & $-$0.004 & 0.000 & $+$0.106 & $+$0.022 \\
    \midrule
    EVA & photographic level & $-$0.127$^\ast$ & $-$0.126$^\ast$ & $+$0.172 & $-$0.113 \\
    EVA & age & $-$0.035$^\ast$ & $-$0.033$^\ast$ & $+$0.246 & $-$0.067 \\
    \bottomrule
  \end{tabular*}
  \caption{Robustness of the group results. The first two columns give the other-half correlation of Table~\ref{tab:validity} for each of the two greedy persona runs, with $^\ast$ marking values that survive the Holm correction over the eight tests. The last two compare a lookup table of group gaps, fitted on real ratings of other raters and other images, with the audience's second run on the same cells, where the audience is still scored on the raters whose cards it received.}
  \label{tab:validity-runs}
\end{table*}

Table~\ref{tab:validity-runs} shows that the two persona runs agree closely, since every other-half correlation moves by at most 0.01 and the same four tests survive the correction. For interest in art the ratio to the uncentered split-half reference, which assumes independent half errors, is 0.83 in the first run against 0.87 in the second, and both EVA tests stay negative. Both runs decode greedily, yet on matched rater and image pairs their overall scores differ on 13\% of PARA, 12\% of EVA and 21\% of LAPIS ratings, by a mean of 0.07, 0.13 and 2.0 points on the raw scales. We attribute these differences to non-deterministic numerics in batched bfloat16 inference, which we have not isolated, so this shows stability across reruns rather than across decoding settings. Both the lookup and the audience use human ratings, the lookup to measure the gaps and the audience to calibrate. The lookup is within 0.012 of the audience on LAPIS interest in art and nationality and on PARA art experience, higher by 0.08 or more on PARA photography experience and age and on both EVA tests, and lower only on LAPIS age, where the audience is scored on the raters it was given and its other-half value is about zero.

\begin{figure}[!htb]
  \centering
  \includegraphics[width=\columnwidth]{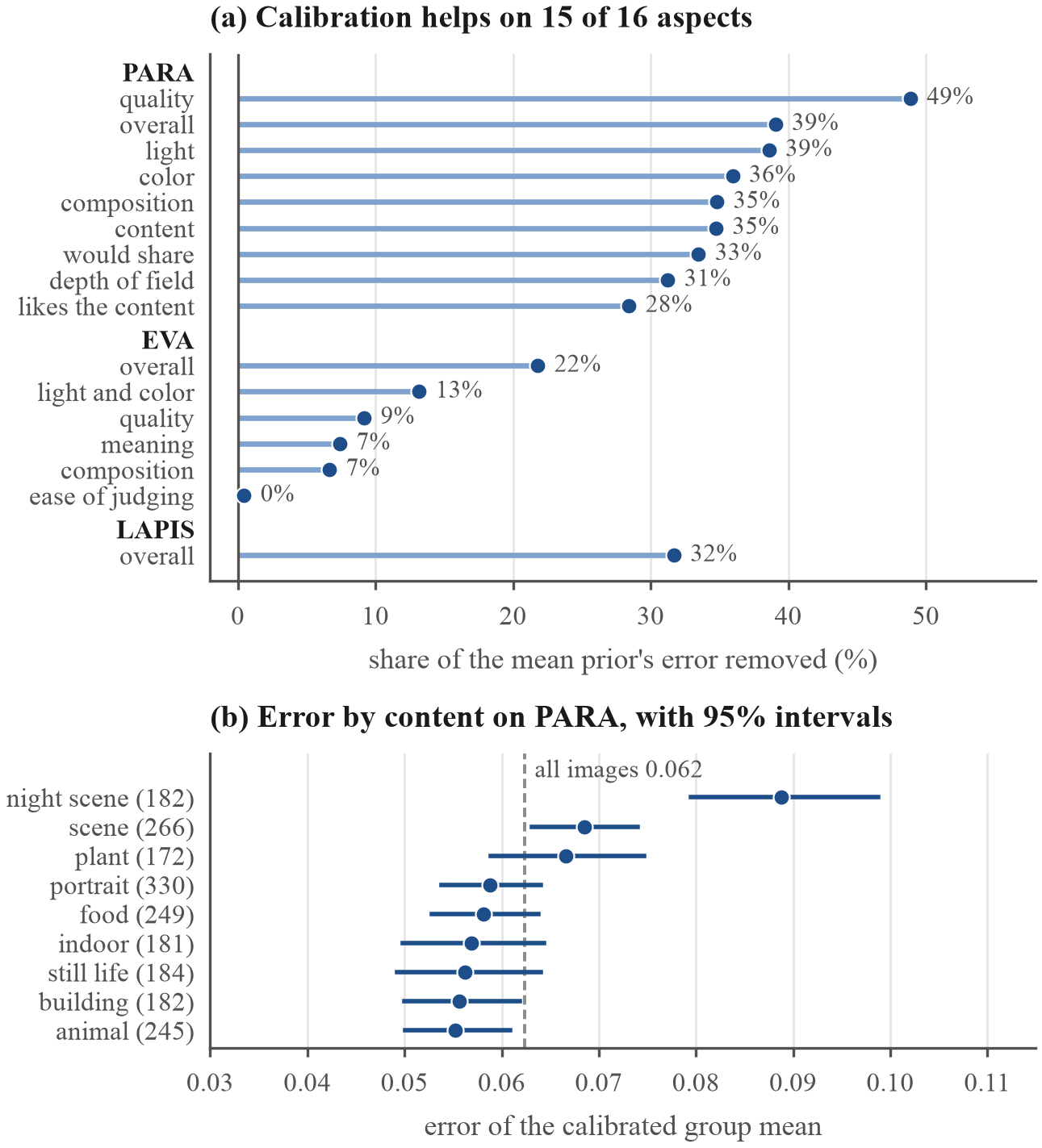}
  \caption{Where the calibrated audience improves over a constant prior, in the first persona run. (a) The share of the prior's error that the calibrated audience removes, for each of the 16 rated aspects. This does not isolate calibration or the personas (see Table~\ref{tab:aspects}). (b) Error of the calibrated audience by image content on PARA, with the number of images and 95\% intervals over images, against the overall error (dashed).}
  \label{fig:breadth}
\end{figure}

Figure~\ref{fig:breadth} shows that the calibrated audience beats the prior on 15 of the 16 rated aspects, not only on the overall score, removing 7 to 49\% of its error, with nothing gained only on the ease of judging. Its error also depends on what the image shows, from 0.055 for animals to 0.089~\ci{0.079}{0.099} for night scenes, whose interval lies above that of every other kind of content.

Table~\ref{tab:aspects} adds the comparison that the figure lacks, on the second persona run and the persona-free panel, calibrated on the same image halves, with a prior fitted on the training half of the images only. The persona-free panel is more accurate on 9 of the 16 aspects, with intervals that exclude zero, including how much the viewer likes the content and would share the photograph, the two aspects that ask most directly about the person. It is not detectably different on 5, and the personas lead only on LAPIS and, by 0.002, on PARA composition. These aspect contrasts were not planned in advance, so we report all of them without significance marks. The overall rows of Table~\ref{tab:aspects} are the values of Table~\ref{tab:calibration}, from the same matched records, image halves and bootstrap, and a prior fitted on all images instead of the training half changes its error by at most 0.0001.

\begin{table*}[t]
  \centering
  \footnotesize
  \begin{tabular*}{\textwidth}{@{\extracolsep{\fill}}llccccc@{}}
    \toprule
    & & \multicolumn{2}{c}{Audience $\downarrow$} & \multicolumn{2}{c}{Persona-free $\downarrow$} & \\
    \cmidrule(lr){3-4}\cmidrule(lr){5-6}
    & Aspect & raw & cal. & raw & cal. & \hd{Calibrated difference\\ {[95\% CI]}} \\
    \midrule
    PARA & aesthetic & 0.182 & 0.062 & 0.127 & 0.056 & $+$0.0061 [0.0044,\,0.0079] \\
     & quality & 0.239 & 0.051 & 0.186 & 0.052 & $-$0.0008 [$-$0.0021,\,0.0007] \\
     & composition & 0.151 & 0.057 & 0.146 & 0.059 & $-$0.0018 [$-$0.0031,\,$-$0.0005] \\
     & color & 0.294 & 0.061 & 0.208 & 0.061 & $+$0.0000 [$-$0.0018,\,0.0019] \\
     & depth of field & 0.275 & 0.064 & 0.207 & 0.062 & $+$0.0021 [0.0004,\,0.0038] \\
     & content & 0.200 & 0.056 & 0.171 & 0.055 & $+$0.0008 [$-$0.0006,\,0.0022] \\
     & light & 0.313 & 0.065 & 0.202 & 0.062 & $+$0.0029 [0.0010,\,0.0047] \\
     & likes content & 0.195 & 0.073 & 0.165 & 0.067 & $+$0.0061 [0.0041,\,0.0079] \\
     & would share & 0.209 & 0.075 & 0.182 & 0.068 & $+$0.0069 [0.0049,\,0.0087] \\
    \midrule
    EVA & overall & 0.110 & 0.063 & 0.124 & 0.061 & $+$0.0026 [0.0017,\,0.0034] \\
     & light and color & 0.309 & 0.077 & 0.297 & 0.077 & $-$0.0002 [$-$0.0009,\,0.0005] \\
     & composition & 0.116 & 0.073 & 0.177 & 0.068 & $+$0.0050 [0.0040,\,0.0059] \\
     & quality & 0.134 & 0.062 & 0.277 & 0.059 & $+$0.0029 [0.0019,\,0.0039] \\
     & content & 0.127 & 0.071 & 0.193 & 0.067 & $+$0.0047 [0.0037,\,0.0058] \\
     & ease of judging & 0.131 & 0.044 & 0.188 & 0.044 & $-$0.0001 [$-$0.0001,\,0.0000] \\
    \midrule
    LAPIS & overall & 0.161 & 0.073 & 0.174 & 0.079 & $-$0.0065 [$-$0.0076,\,$-$0.0053] \\
    \bottomrule
  \end{tabular*}
  \caption{Every rated aspect, as the mean absolute error of each image's mean rating on the $[0,1]$ scale, in the second persona run and the persona-free panel on shared ratings and the same image halves. The difference is calibrated audience minus calibrated persona-free per image, so a negative value favors the personas, with a 95\% interval over images. The training-half prior errors are 0.085 to 0.112 on PARA, 0.044 to 0.088 on EVA and 0.105 on LAPIS.}
  \label{tab:aspects}
\end{table*}

\begin{table*}[t]
  \centering
  \footnotesize
  \begin{tabularx}{\textwidth}{ll>{\raggedright\arraybackslash}Xc}
    \toprule
    Check & Dataset & What we measure & Value \\
    \midrule
    Artist fame      & LAPIS & rank correlation of an artist's number of ratings with the error & $+0.03$ \\
    Painting style   & LAPIS & error gap between the easiest and the hardest style & $0.04$ \\
    Repeated ratings & LAPIS & group error under each of three ways of handling repeats & $0.072$ \\
    Rater agreement  & LAPIS & correlation of rater disagreement with the group error & $-0.19$ \\
    Stated difficulty & EVA  & correlation of the VLM's stated difficulty with rater disagreement & $+0.14$ \\
    Linked aspects   & EVA   & agreement of the VLM's links between rated aspects with people's & $0.77$ \\
    Linked aspects   & PARA  & the same on the nine rated aspects of PARA & $0.29$ \\
    \bottomrule
  \end{tabularx}
  \caption{Further indirect checks related to memory and bookkeeping, each on the datasets where it can be run. They are of limited diagnostic purpose and cannot rule out that the model saw these images or ratings in pretraining. The VLM links the rated aspects much as people do on EVA but far less so on PARA.}
  \label{tab:robust}
\end{table*}

Table~\ref{tab:robust} gives indirect checks, for example that artists rated more often are not predicted better. The prompt has no example images, so the overlap between EVA and AVA cannot leak ratings through the prompt, but these checks cannot establish whether pretraining exposed the model to the images or their ratings, which remains unresolved.

\section{Who the Audience Speaks For}
\label{sup:fairness}
We check whether the VLM treats all kinds of viewers alike, since a tilt toward viewers described one way would pass unseen into every average built on it. For each level of each attribute of the persona cards, such as one nationality, we measure the VLM's bias after calibration, the mean of its calibrated rating minus the human rating. A level that holds few raters inherits the taste of those few, which the tested profile regressions do not predict (Section~\ref{sec:ceiling}), so we compare each level with the range that rater-to-rater noise alone gives a level of its size, the 95\% funnel of Figure~\ref{fig:bias}.

\begin{figure}[!htb]
  \centering
  \includegraphics[width=\columnwidth]{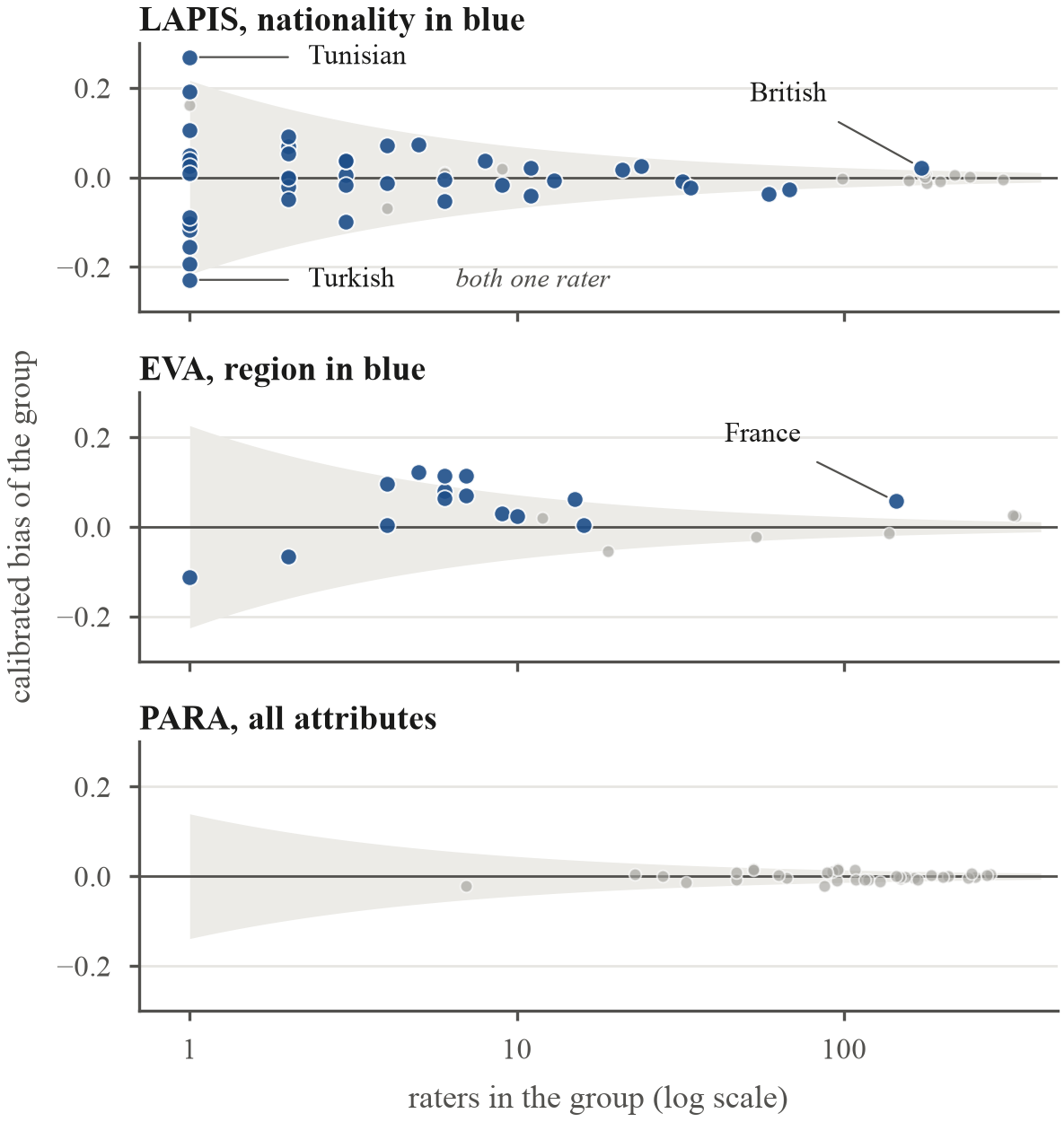}
  \caption{The VLM's bias after calibration for every level of every persona attribute, against the number of raters in that level on a log scale. The shaded funnel is the 95\% range that differences between individual raters produce in a level of that size when the attribute has no effect. Nationality on LAPIS and region on EVA are in blue and all other attributes in gray. The most extreme nationalities are single raters.}
  \label{fig:bias}
\end{figure}

Figure~\ref{fig:bias} shows that the largest differences come from the smallest groups. On LAPIS, the most and least favored nationalities, Tunisian at $+0.27$ and Turkish at $-0.23$, are each a single rater, so the spread of 0.50 between them measures two people's taste rather than their nationality. Among the seven nationalities with at least 20 raters the bias differs by 0.06, although three large ones, British, Polish and South African, still fall outside the descriptive reference band, by at most 0.04. Overall, 5 of the 42 nationalities fall outside the funnel, where chance alone would place about two. On EVA, 5 of 16 regions fall outside it, with biases from 0.06 for France, with 144 raters, to 0.12 for regions of 5 to 7 raters, and on PARA every level of every attribute stays within 0.03 of zero. The funnel treats raters within a level as independent and ignores the image clustering and the calibration fit, so we read it as a descriptive reference rather than a test. The VLM may therefore treat viewers from some places somewhat differently, but far less than a range across single raters suggests. We make no claim that the audience distinguishes viewers by country, and across rater halves it reproduces almost none of the real difference between nationalities on LAPIS (Section~\ref{sec:audience}).

\section{The Editing Loop in Detail}
\label{sup:loop}
Figure~\ref{fig:loop} shows the loop that Section~\ref{sec:c4} uses as a downstream test of the audience. We refine 100 photographs, 50 from PARA and 50 from EVA, for ten rounds each. They are drawn from the photographs whose mean human rating lies between the 15th and the 60th percentile of their dataset, so that each has room to improve, and they are taken in a fixed order that keeps every smaller set nested in a larger one.

\begin{figure*}[t]
  \centering
  \begin{tikzpicture}[
    font=\footnotesize,
    stage/.style={rectangle, rounded corners=3pt, draw=black!60, line width=0.6pt, fill=white,
                  align=center, inner sep=3pt, minimum height=10mm},
    aud/.style={stage, draw=aesyn, fill=aesyn!6,
                double copy shadow={shadow xshift=0.6ex, shadow yshift=0.6ex,
                                    draw=aesyn!60, fill=aesyn!3}},
    note/.style={font=\scriptsize\itshape, text=black!60},
    ringnote/.style={font=\scriptsize\itshape, text=aesyn, inner sep=1pt},
    spoke/.style={-{Latex[length=2.2mm]}, line width=0.7pt, draw=black!70},
    ring/.style={line width=1.1pt, draw=aesyn, postaction={decorate}, decoration={markings,
                 mark=at position 0.0615 with {\arrow{Latex[length=2.6mm]}},
                 mark=at position 0.25 with {\arrow{Latex[length=2.6mm]}},
                 mark=at position 0.4485 with {\arrow{Latex[length=2.6mm]}},
                 mark=at position 0.5963 with {\arrow{Latex[length=2.6mm]}},
                 mark=at position 0.9037 with {\arrow{Latex[length=2.6mm]}}}},
  ]
    \draw[ring] (-5.6,0) arc[start angle=180, end angle=-180, x radius=5.6, y radius=2.5];

    \node[stage, fill=black!3] (cur) at (-5.6,0)
      {\includegraphics[width=0.8cm]{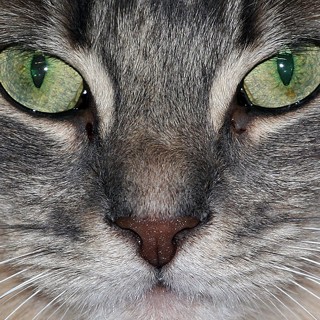}\\[1pt]Current best image\\[-1pt]{\scriptsize the source at first}};
    \node[aud] (crit) at (-3.212,2.048)
      {Audience critique\\[-1pt]{\scriptsize ten PARA persona cards}\\[-2pt]{\scriptsize role-played by \qwenmark\,Qwen2-VL-7B}};
    \node[stage] (mrg) at (3.212,2.048)
      {Merge into one\\[-1pt]instruction\\[-1pt]{\scriptsize Qwen2-VL-7B, at most 15 words}};
    \node[stage] (ed) at (5.6,0)
      {Image editor\\[-1pt]{\scriptsize \fluxmark\,FLUX.1-Kontext-dev}\\[-2pt]{\scriptsize three seeds per round}};
    \node[stage] (sc) at (0,-2.5)
      {Selection scorer\\[-1pt]{\scriptsize\raisebox{-0.3ex}{\includegraphics[height=5pt]{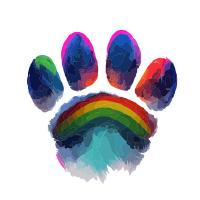}}\,LAION aesthetic score}\\[-2pt]{\scriptsize \dinomark\,DINOv2 similarity $\geq 0.78$}};

    \draw[spoke] ($(mrg.north)+(-0.45,0)$) -- ++(0,0.32) -- ++(0.9,0) -- ++(0,-0.32);
    \node[note, anchor=south] at ($(mrg.north)+(0,0.34)$) {previous instruction};

    \node[inner sep=0, draw=black, line width=0.5pt] (orig) at (0,0.3)
      {\includegraphics[width=1.05cm]{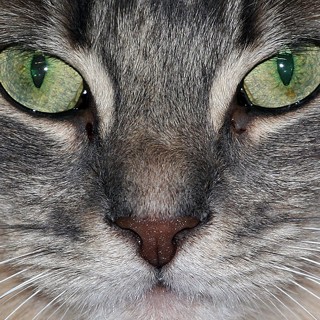}};
    \node[font=\footnotesize, anchor=north, inner sep=2pt] (origlab) at (orig.south) {original image};
    \draw[spoke] (orig.east) -- node[note, above]{every edit starts here} (orig.east -| ed.west);
    \draw[spoke, dashed, draw=black!55] (origlab.south) -- node[note, right]{similarity} (sc.north);
    \node[ringnote] at (0,1.55) {10 rounds};

    \node[ringnote, anchor=south east] at (-5.18,1.4) {seen by every persona};
    \node[ringnote, anchor=south] at (0,2.75) {ten complaints};
    \node[ringnote, anchor=south west] at (5.18,1.4) {one instruction};
    \node[ringnote, anchor=north west] at (4.09,-2.06) {three candidates};
    \node[ringnote, anchor=north east, align=right] at (-4.09,-2.06) {kept only if it beats\\the current best};
  \end{tikzpicture}
  \caption{The editing loop with the audience as critic, drawn as the cycle of Figure~\ref{fig:overview}c. Each round runs clockwise from the current best image, which is the source until an edit is kept. The ten personas each name one change, the same model merges them with the previous instruction, FLUX.1-Kontext-dev applies the result to the original image with three seeds, and the scorer keeps the best candidate with DINOv2 similarity of at least 0.78 if it beats the current best.}
  \label{fig:loop}
\end{figure*}

\subsubsection{One Round.} Each round has four steps. First, each of the ten personas looks at the current best image, which is the source until an edit is kept, and returns a score from 0 to 100 and the single change it would most like, under the PARA system prompt of the synthetic audience with its own card and the critic question of Appendix~\ref{sup:prompts}. The ten personas are PARA raters drawn once at random and kept for every image, including the EVA photographs, and we record their scores but never use them. Second, the same model, which also sees the current best image, merges the current instruction with the new complaints into one instruction of at most fifteen words. It does so in every round, whether or not the last edit was kept, so requests accumulate across rounds and a rejected request can stay in the instruction until a later merge drops it. Third, FLUX.1-Kontext-dev \cite{flux2024} applies the instruction, followed by the request for a visible edit, to the original image rather than to the current best, so that editing artifacts do not compound, with three seeds per round ($1000r$ to $1000r+2$ in round $r$, the same for every critic), a guidance of 3.0 and 28 sampling steps. Fourth, among the candidates whose similarity to the original, the cosine between the DINOv2-base \cite{oquab2023} class-token embeddings of the two images, is at least 0.78, the loop takes the one that the selection scorer rates highest and keeps it only if it scores higher than the best image so far. Since every candidate is a single edit of the original, a kept edit replaces the previous one rather than adding to it, so the final image is the best of up to 30 single edits, and an edit kept in an early round can be replaced later, as in the worked example below.

\subsubsection{The Four Critics.} The audience gathers one complaint from each of its ten personas. The persona-free critic makes one call per round under the persona-free PARA system prompt, and its critique is merged in the same way. The fixed instruction repeats one fixed instruction to improve the image, and the generic-instruction cycle (\pt{reward\_only} in the run logs) cycles through eight generic instructions, so that only selection guides it, and neither of these two looks at the image or uses the merge. The critics and the merge decode greedily, with up to 96 and 48 new tokens, so the audience makes 110 model calls per image, 100 critiques and ten merges, and the persona-free critic 20. Every critic appends the same request for a visible edit, and we do not score whether an edit follows its instruction, which would favor whichever critic writes the easiest instruction.

\subsubsection{The Selection Scorer.} The LAION aesthetic predictor \cite{schuhmann2022predictor} embeds an image with a frozen CLIP ViT-L/14 encoder, normalizes the embedding to unit length and maps it to one number with a small head fitted to human ratings of generated images, AVA photographs and logos, so it estimates the average rating these raters would give, on a scale of about 1 to 10. All editing scores and gains in this paper are on this scale, not on the 0 to 1 scale of the rating results. The predictor reads only the image and is not conditioned on any viewer, and since it both selects the edits and measures the gain, the gains are not an outside measure of quality. The EVA photographs also come from AVA, one of the datasets on which the predictor was trained \cite{schuhmann2022}. We set the similarity floor of 0.78 by hand, low enough that visible edits pass and high enough that the photograph stays recognizable.

\begin{table*}[t]
  \centering
  \footnotesize
  \setlength{\tabcolsep}{3pt}
  \begin{tabular*}{\textwidth}{@{\extracolsep{\fill}}lcccccccc@{}}
    \toprule
    & \multicolumn{5}{c}{Selection scorer} & \multicolumn{2}{c}{Instructions} & \\
    \cmidrule(lr){2-6}\cmidrule(lr){7-8}
    Critic & gain $\uparrow$ [95\% CI] & \hd{images\\improved $\uparrow$} & \hd{mean candidate\\change [95\% CI]} & \hd{first-round\\candidates} & \hd{gate\\pass} & \hd{distinct\\per image} & \hd{content edits\\kept} & \hd{VLM calls\\per image $\downarrow$} \\
    \midrule
    fixed instruction & 0.164 [0.134,\,0.195] & 84\% & $+$0.010 [$-$0.021,\,0.042] & $+$0.012 & 100\% & 1.00 & 0\% & 0 \\
    persona-free critic & 0.155 [0.127,\,0.185] & 72\% & $-$0.044 [$-$0.083,\,$-$0.003] & $-$0.017 & 90\% & 2.83 & 25\% & 20 \\
    audience & 0.170 [0.136,\,0.207] & 69\% & $-$0.025 [$-$0.062,\,0.014] & $-$0.007 & 85\% & 3.14 & 44\% & 110 \\
    \midrule
    \textit{generic cycle} & 0.187 [0.157,\,0.220] & 86\% & $+$0.017 [$-$0.014,\,0.048] & $+$0.012 & 100\% & 8.00 & 2\% & 0 \\
    \bottomrule
  \end{tabular*}
  \caption{Each critic over ten rounds on 100 photographs, scored by the selector that also chooses the edits. Mean candidate change averages all 30 logged candidates of an image along the adaptive trajectory, and first-round averages the three candidates of round one. Gate pass is the share of candidates with DINOv2 similarity of at least 0.78. Content edits kept is the share of rounds with a kept edit whose instruction a keyword tagger labels as adding, removing or replacing an object, out of 213, 163, 171 and 208 such rounds (75 of 171 for the audience). Generic cycle is the generic-instruction cycle. Every other quantity weights images equally, with intervals over images.}
  \label{tab:editing}
\end{table*}

Table~\ref{tab:editing} gives the full comparison. Besides the results in Section~\ref{sec:c4}, the persona-free critic's candidates average $-0.044$, the only candidate mean whose interval excludes zero, and the generic-instruction cycle's $+0.017$. These means describe the logged trajectories, whose later instructions depend on what was selected, and the first-round means, which precede any selection, are all within 0.03 of each other. Only 47\% of the audience's candidates score above the source, against 53\% for the fixed instruction, and the audience improves fewer images, 69 against 84\%. The gate removes 15\% of the audience's candidates and 10\% of the persona-free critic's but almost none of the fixed critics', and 116 of the audience's 1{,}000 rounds and 72 of the persona-free critic's have no eligible candidate, in which case the loop keeps its current best image. Seven of the audience's images and four of the persona-free critic's have no eligible candidate in any round and keep the source, and they stay in every comparison with a gain of zero. Among eligible candidates alone, a mean conditional on the gate and defined for 93 of the audience's images, the audience's candidates average $-0.024$. Paired over images at 30 candidates, the audience minus the generic-instruction cycle is $-0.017$~\ci{$-0.044$}{0.014}, and after 15 candidates it is $-0.030$~\ci{$-0.057$}{0.001}. The audience writes 3.1 distinct instructions per image and repeats the previous one in 75\% of rounds, as the persona-free critic does. Each condition ran once with the same seed schedule, three editor seeds per round that change from round to round and are shared by all critics, so the intervals cover variation over images and not over runs.

\subsubsection{A Worked Example.} Figure~\ref{fig:loop-example} follows the cat of Figure~\ref{fig:overview} through the audience's loop, and the boxes below give what its ten personas asked for in the first three rounds. The ten cards describe six women and four men aged 18 to 40, with art experience from beginner to expert. In round one all ten give the source 80 out of 100 and ask for three changes, which the merge joins into one instruction. Of the three candidates, the one that changes the image most, at a similarity of 0.87, scores 5.522 against 5.307 for the source, so the loop keeps it. In round two the personas look at that edit, all give it 70 and all ask to remove the green eyes, and the loop keeps a candidate that paints flat cartoon eyes on the cat, because it scores 5.545, 0.023 above the round-one edit. In round three all ten ask to remove the whiskers to make the image more abstract, and no candidate scores higher. The instruction then stays the same until round ten, and in round seven one of its candidates, which leaves the eyes almost as they were, scores 5.560 and replaces the cartoon eyes, so the audience's final image in Figure~\ref{fig:c4-worst5} barely differs from the source. The fixed instruction's first edit, which mainly sharpens the fur, scores 5.83, and no critic does better on this photograph. Ten different cards thus produced at most three distinct requests per round, two of which would strip a cat of its whiskers and its green eyes, and the scorer kept the second for a gain of 0.023 on its scale of about 1 to 10.

\begin{instrbox}{Round 1, complaints of the ten personas (each scored the source 80 out of 100)}
$\bullet$~remove the whiskers to make the image more symmetrical \hfill (5 personas)\\
$\bullet$~blur the background to make the cat's face stand out more \hfill (3 personas)\\
$\bullet$~desaturate the background to better highlight the cat's fur and eyes \hfill (2 personas)
\end{instrbox}
\begin{instrbox}{Round 1, instruction for the editor (kept, score 5.307 to 5.522)}
\itshape Remove whiskers, blur background, desaturate background, focus on cat's face.
\end{instrbox}
\begin{instrbox}{Round 2, complaints (each scored 70) and instruction (kept, 5.522 to 5.545)}
$\bullet$~remove the green eyes to make the cat look more natural \hfill (10 personas)\\[2pt]
\itshape Remove green eyes and make cat's eyes normal.
\end{instrbox}
\begin{instrbox}{Round 3, complaints (eight scored 80, two 70) and instruction (not kept)}
$\bullet$~remove the whiskers to make the image more abstract and focus on the eyes \hfill (10 personas)\\[2pt]
\itshape Remove whiskers and adjust eyes to normal appearance.
\end{instrbox}

\begin{figure*}[t]
  \centering
  \includegraphics[width=\textwidth]{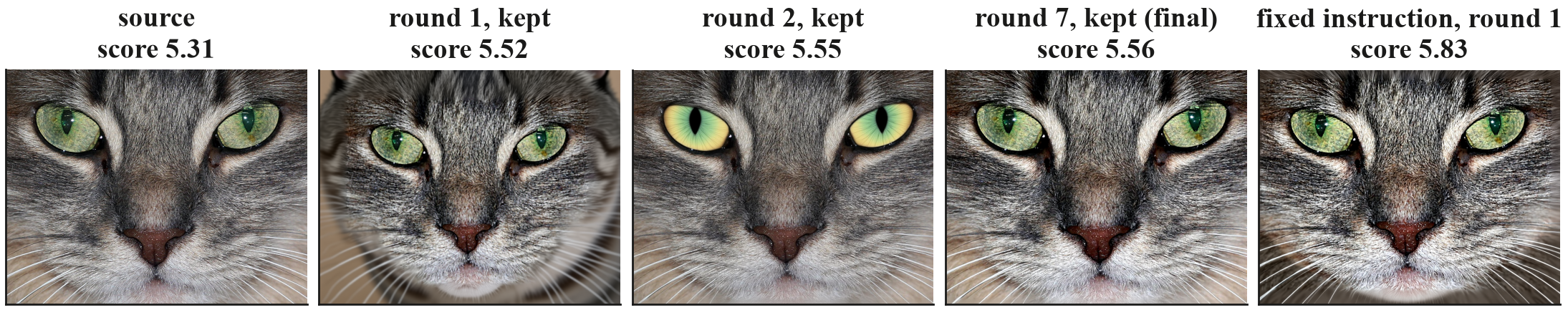}
  \caption{The cat of Figure~\ref{fig:overview} under the audience's loop, with the selection score of each image. From the left, the source, the edit kept in round one, the edit kept in round two, which Figure~\ref{fig:overview} shows, and the edit kept in round seven, which replaces it and is the audience's final image in Figure~\ref{fig:c4-worst5}, with the fixed instruction's first edit, which no critic beats on this photograph, on the right. All five scores come from the loop's own selector.}
  \label{fig:loop-example}
\end{figure*}

\section{Editing Examples}
\label{sup:c4qual}
Figure~\ref{fig:examples-main} shows each critic's final image on three photographs where the audience keeps a visible edit but a baseline ends higher. Per image, the fixed instruction ends above the audience on 49 of the 100 images and below it on 40, the generic-instruction cycle above it on 53 and below it on 36, and the persona-free critic above it on 34 and below it on 44, with ties on the rest.

\begin{figure*}[t]
  \centering
  \includegraphics[width=\textwidth]{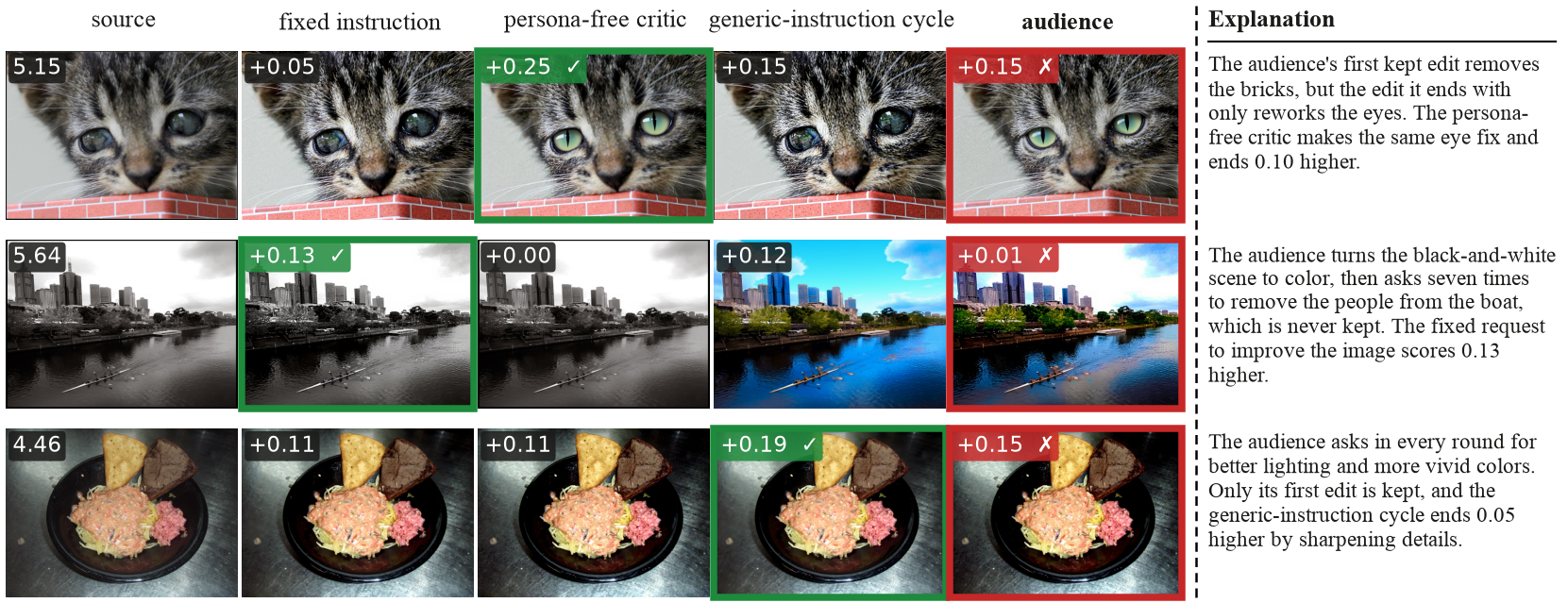}
  \caption{Each critic's final image on three photographs where the audience keeps a visible edit but a baseline ends higher. Badges give the source's selection score and each critic's change from it, green marks the highest score and red the audience, and the right column summarizes the run logs. We chose the first two rows, and the third is drawn at random (more random rows are in Figure~\ref{fig:c4-random}).}
  \label{fig:examples-main}
\end{figure*}

Figure~\ref{fig:c4-random} shows the final image of every critic for three photographs drawn at random from the 100 of the editing loop (seed 0). We show unselected examples, since images chosen for a large gain would not show a typical edit. The edits are small, and on every row the four critics end within 0.09 of each other. The audience ends highest on none of the three rows, and on the glass door it asks ten times to remove the reflections, every candidate scores below the source, and it keeps no edit at all.

\begin{figure*}[p]
  \centering
  \includegraphics[width=\textwidth]{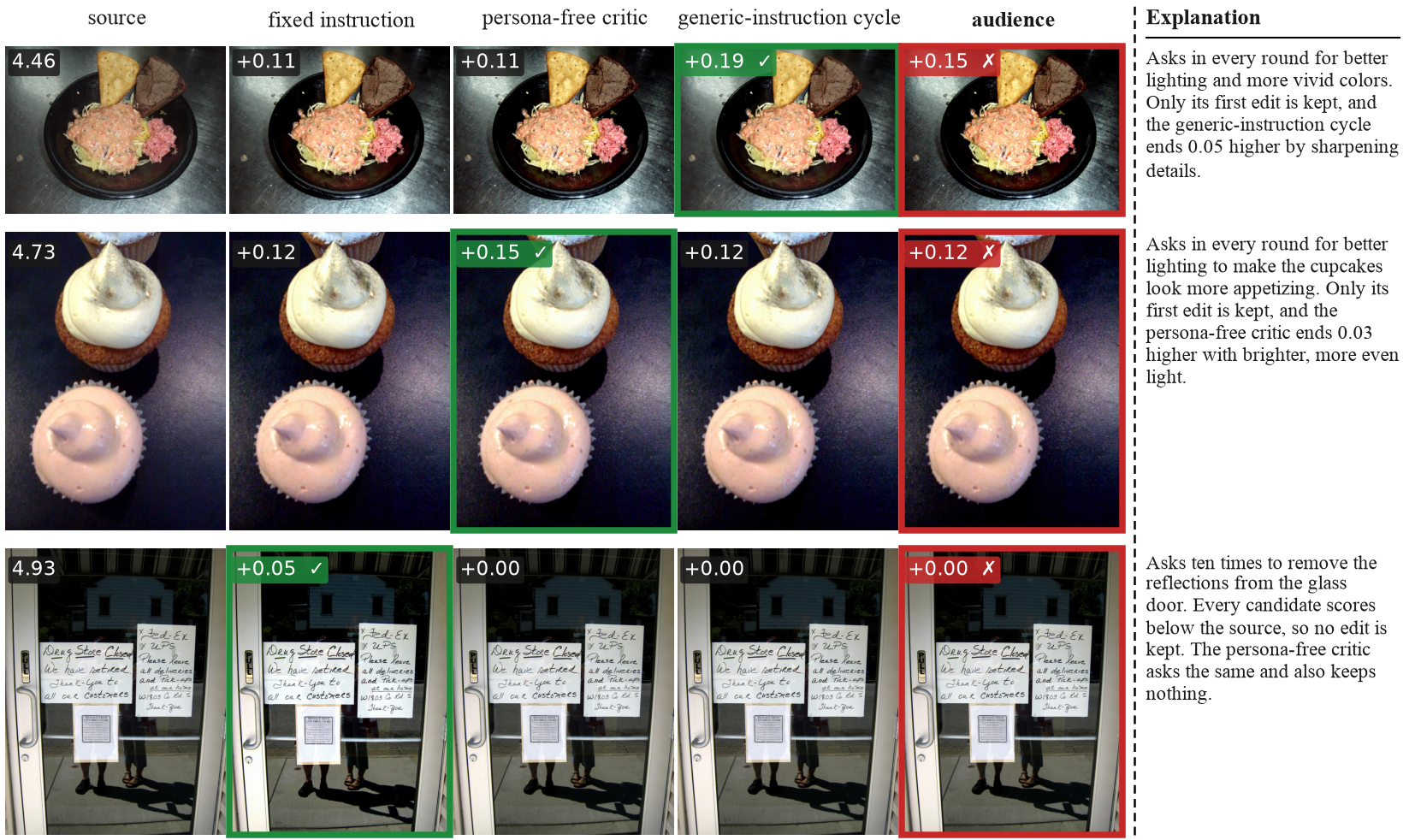}
  \caption{Each critic's final image for three photographs drawn at random from the 100 of the editing loop. Badges, check marks, crosses and frames are as in Figure~\ref{fig:examples-main}, and the right column describes, from the run logs, what the audience asked for and what the loop kept. The edits are small and the critics' scores are close, as in Table~\ref{tab:editing}.}
  \label{fig:c4-random}
\end{figure*}

\begin{figure*}[p]
  \centering
  \includegraphics[width=\textwidth]{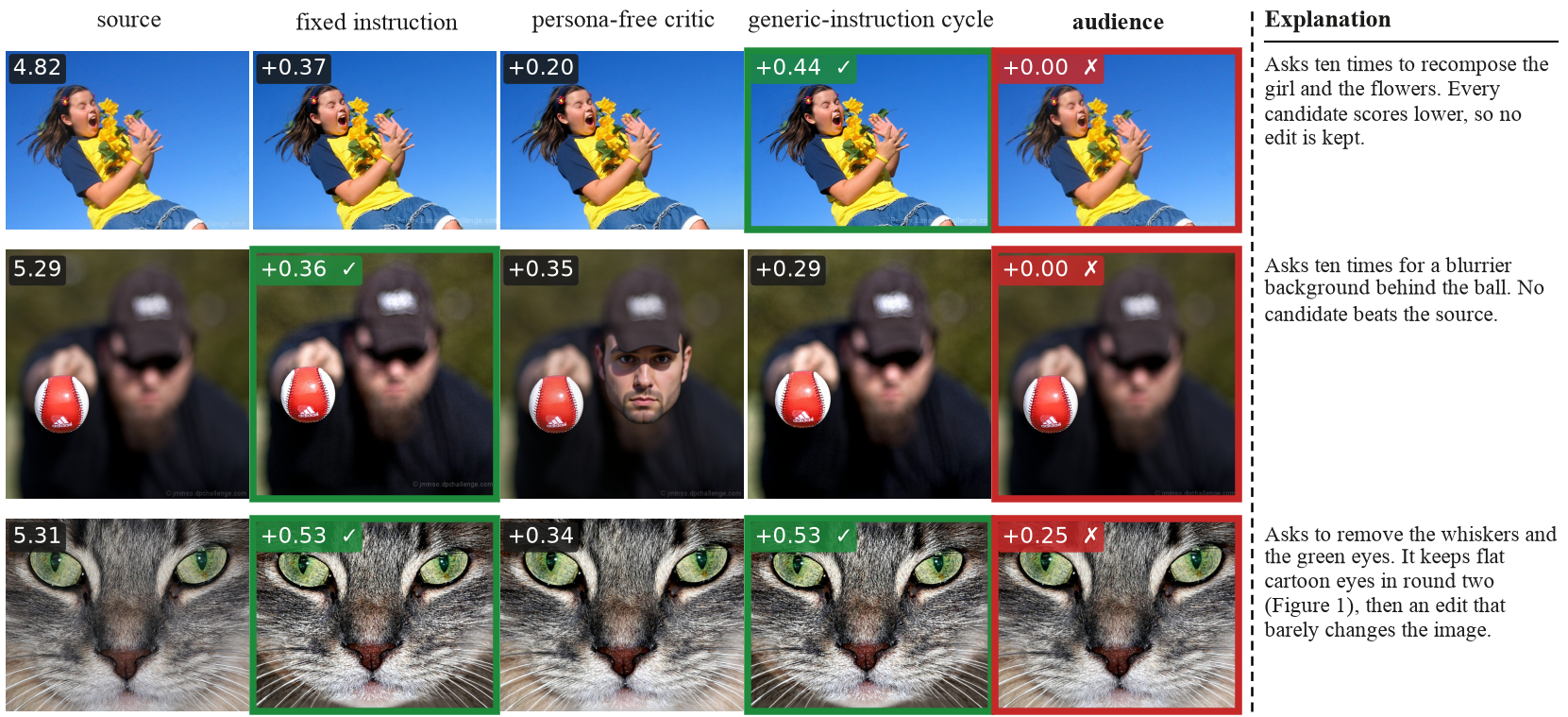}
  \caption{Each critic's final image for the three images on which the audience ends furthest below the fixed instruction, chosen for that reason, shown as in Figure~\ref{fig:c4-random}.}
  \label{fig:c4-worst5}
\end{figure*}

Figures~\ref{fig:c4-worst5} and~\ref{fig:c4-worst-more} show the opposite of a random draw, the 11 images on which the audience ends furthest below the fixed instruction. The right column describes, from the run logs, what the audience asked for and why the loop kept little or nothing. Three failures recur. The audience repeats one request for all ten rounds although every candidate scores lower, it asks to remove or replace objects, which pushes every edit below the similarity floor, and it piles up requests until no later edit is kept. On the same images the fixed instruction and the generic-instruction cycle gain with plain changes to light and color.

\begin{figure*}[p]
  \centering
  \includegraphics[width=\textwidth]{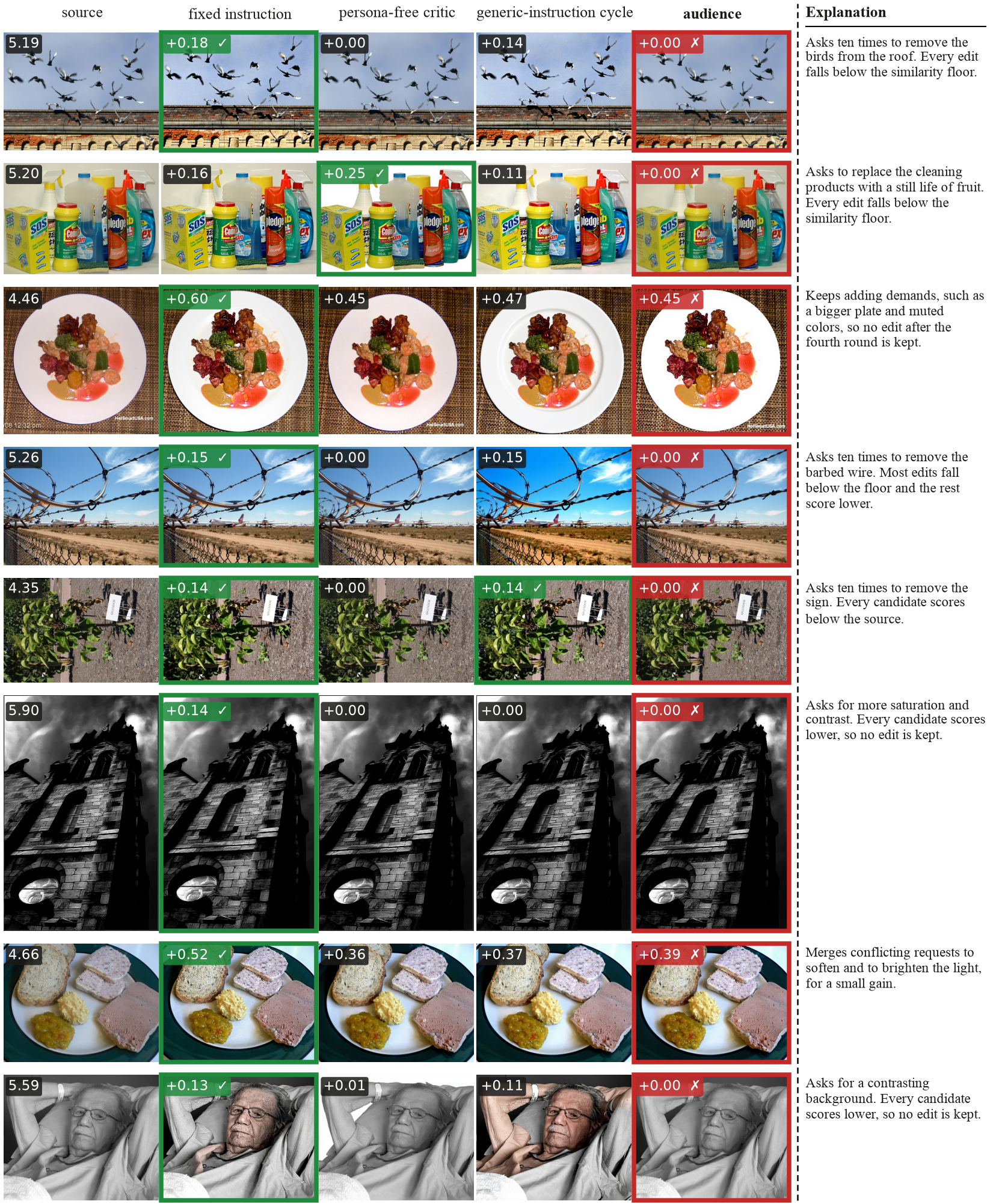}
  \caption{The next eight images on which the audience ends furthest below the fixed instruction, shown as in Figure~\ref{fig:c4-random}.}
  \label{fig:c4-worst-more}
\end{figure*}

Figure~\ref{fig:c4-success} shows the other side of Figure~\ref{fig:examples-main}, six images chosen by hand where the editing loop with the audience as critic does end highest. The audience alone ends highest on 22 of the 100 images, and we picked these rows from those 22 for a visible kept edit and varied content, so they show what its critique can do at best and not what it usually does. On four of the six rows the audience wins by asking to add or replace objects, such as a player behind a basketball, a truck in an empty courtyard or a Milky Way over a dark road, which the fixed instruction and the generic-instruction cycle never ask for, and on the other two it asks to remove an oven or a busy background. The Milky Way gains 1.06, the largest gain of any critic on the 100 images, while the baselines gain at most 0.21 on the same photo.

\begin{figure*}[p]
  \centering
  \includegraphics[width=\textwidth]{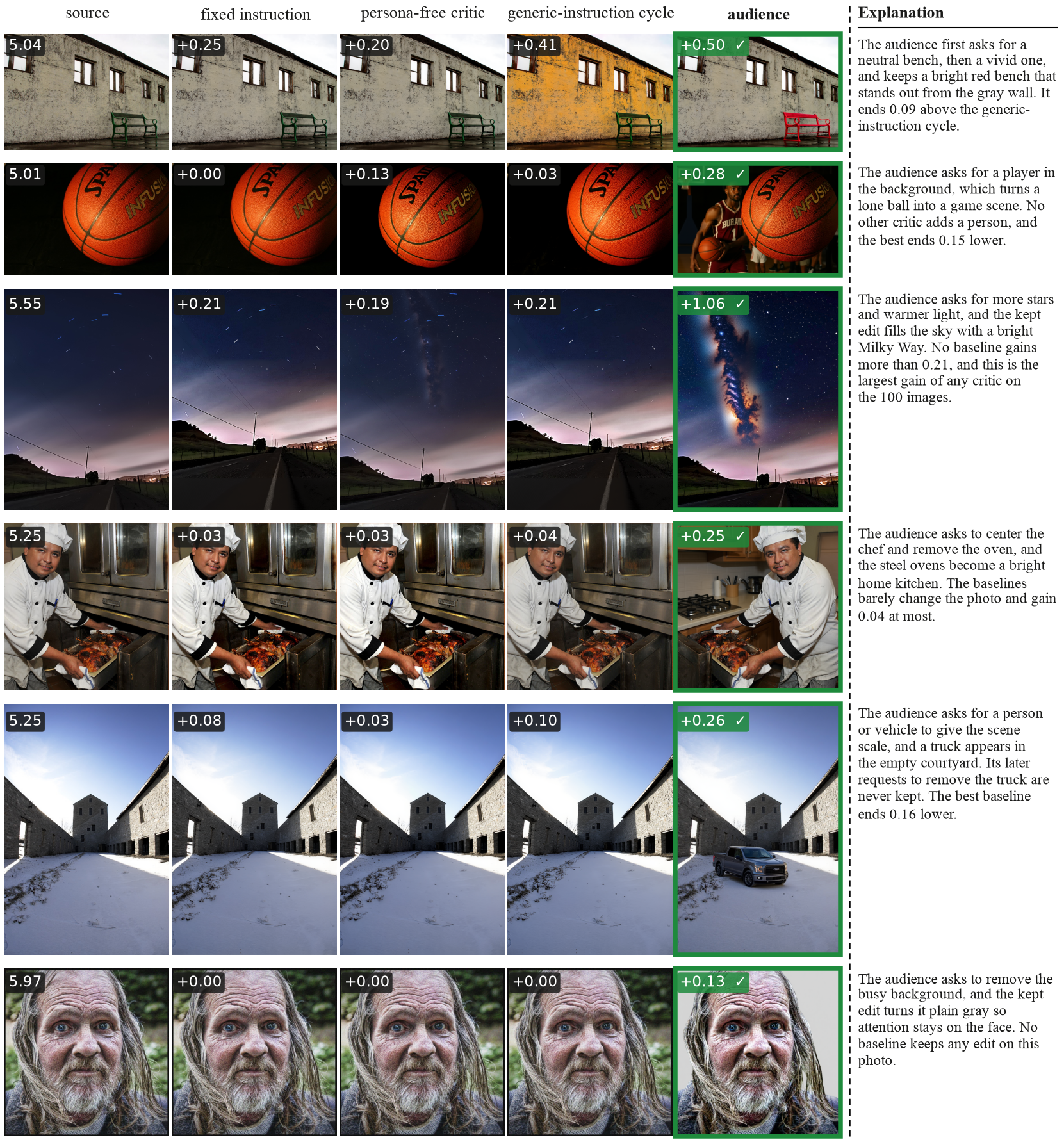}
  \caption{Selected examples where the audience's edits reach the highest score, the counterpart of Figure~\ref{fig:examples-main}. We chose these six rows by hand from the 22 images on which the audience alone ends highest, so they are not typical. Badges and frames are as in Figure~\ref{fig:examples-main}, and the right column summarizes the run logs.}
  \label{fig:c4-success}
\end{figure*}

\section{Prompts}
\label{sup:prompts}
Every judgment in this paper comes from the frozen VLM answering a system prompt and a question about an image. No generative-model weights are updated, and only calibration maps and diagnostic predictors are fitted. The prompts below are reproduced as they generated the saved outputs, without later edits. Each experiment has its own prompts. The rating experiments on PARA, EVA and LAPIS in Sections~\ref{sec:generated} to~\ref{sec:persona} and Appendices~\ref{sup:calib} to~\ref{sup:fairness} use the rating prompt of that dataset, whose question asks for every aspect the dataset collected. The AI-generated pairs of Rapidata, also in Section~\ref{sec:generated}, use a pairwise prompt that asks which of two images the voter prefers. The editing loop of Section~\ref{sec:c4} uses a critic prompt that asks for a score and a single change, and a further prompt that turns the complaints gathered so far into one instruction for the image editor. In the rating experiments and the editing loop, the synthetic audience and the persona-free panel receive the same task and differ only in the system prompt and in whether the question asks for an answer in character. The persona card is inserted verbatim at \pt{\{description\}}.

\subsubsection{Rating Prompts (PARA, EVA, LAPIS).} Each simulated rating is one call, in which the system prompt below introduces the rater through their persona card and the multi-aspect question further down asks for the rating. The three system prompts follow one template and differ in how they describe the study and which attributes of the rater they name, and the LAPIS prompt adds a sentence on how widely taste in art varies.
\begin{promptbox}{PARA system prompt (synthetic audience)}
\ttfamily
You are role-playing as one specific human annotator from a photo-aesthetics rating study. The annotator is \{description\}\\[2pt]
You are shown one photograph from the study. Judge it the way this exact person would: let their age, gender, education, art and photography experience, and Big-Five personality shape both what they notice and how critical or generous they are. Not every photo deserves a good score; be honest when a photo would bore or disappoint this person. Do not mention that you are an AI or that you are role-playing.
\end{promptbox}
\begin{promptbox}{PARA system prompt (persona-free panel)}
\ttfamily
You are an impartial, general-purpose judge of photographic aesthetics.\\[2pt]
You are shown one photograph from a photo-aesthetics rating study. Judge it on its own merits the way a typical viewer would, without adopting any particular person's perspective or taste. Not every photo deserves a good score; be honest when a photo is mediocre or disappointing. Do not mention that you are an AI.
\end{promptbox}
\begin{promptbox}{EVA system prompt (synthetic audience)}
\ttfamily
You are role-playing as one specific human participant in an online study of visual aesthetics. The participant is \{description\}\\[2pt]
You are shown one photograph from the study. Judge it the way this exact person would: let their age, gender, cultural background, photography experience, and eyesight shape both what they notice and how critical or generous they are. Not every photo deserves a good score; be honest when a photo would bore or disappoint this person. Do not mention that you are an AI or that you are role-playing.
\end{promptbox}
\begin{promptbox}{EVA system prompt (persona-free panel)}
\ttfamily
You are an impartial, general-purpose judge of visual aesthetics.\\[2pt]
You are shown one photograph from a study of visual aesthetics. Judge it on its own merits the way a typical viewer would, without adopting any particular person's perspective or taste. Not every photo deserves a good score; be honest when a photo is mediocre or disappointing. Do not mention that you are an AI.
\end{promptbox}
\begin{promptbox}{LAPIS system prompt (synthetic audience)}
\ttfamily
You are role-playing as one specific human participant in an online study of aesthetic appreciation of artworks. The participant is \{description\}\\[2pt]
You are shown one artwork (a painting or drawing) from the study. Judge it the way this exact person would: let their age, nationality, gender, education, and level of interest in art shape both what they notice and how critical or generous they are. People differ widely in taste — some love abstract art, others only figurative work; not every artwork deserves a good score, so be honest when a piece would bore or leave this person cold. Do not mention that you are an AI or that you are role-playing.
\end{promptbox}
\begin{promptbox}{LAPIS system prompt (persona-free panel)}
\ttfamily
You are an impartial, general-purpose judge of artistic aesthetics.\\[2pt]
You are shown one artwork (a painting or drawing) from a study of aesthetic appreciation. Judge it on its own merits the way a typical viewer would, without adopting any particular person's perspective or taste. Tastes differ widely and not every artwork deserves a good score; be honest when a piece is mediocre or unappealing. Do not mention that you are an AI.
\end{promptbox}

\subsubsection{Persona Cards (PARA, EVA, LAPIS).} A card is filled in from the rater's own record with the templates below. Each band is low, moderate or high, with cut-offs at 4 and 7 for the Big-Five scores and at 2 and 4 for the VAIAK art-interest score. On EVA the age is 2020 minus the recorded birth year and the photographic level reads \pt{has no experience with photography}, \pt{is a beginner at photography}, \pt{is an amateur photographer} or \pt{is a professional photographer}. On EVA and LAPIS the clauses on glasses and color blindness appear only for raters who reported them. A field that the rater left blank is left out of the card.
\begin{promptbox}{Persona card templates}
\ttfamily
{\rmfamily\bfseries PARA}\quad a <gender> aged <age band>, education level: <education>; art experience: <level>; photography experience: <level>. Big-Five personality scores (2-10 scale): extraversion <x>/10 (<band>), agreeableness <x>/10 (<band>), neuroticism <x>/10 (<band>), openness to experience <x>/10 (<band>), conscientiousness <x>/10 (<band>).\\[4pt]
{\rmfamily\bfseries EVA}\quad a <gender> person aged <age> from <region>; <photographic level>; wears corrective glasses; is colour-blind.\\[4pt]
{\rmfamily\bfseries LAPIS}\quad a <gender> <nationality> person aged <age>; education level: <education>; interest in art: <band> (VAIAK art-interest score <x> on a 0-6 scale); is colour-blind, but still perceives colours.
\end{promptbox}

\begin{sloppypar}
\subsubsection{Multi-aspect Rating Questions.} The question asks for every aspect the dataset collected in a single JSON answer, nine on PARA, six on EVA and one on LAPIS, which collects only the overall rating. Each aspect takes one line that names it, explains its ends and gives its range and step, so the model answers on the scale the raters used (Section~\ref{sec:method}). The overall score on the first line is the one we study throughout the paper, and the other aspects enter only the breadth, difficulty and linked-aspect checks of Appendix~\ref{sup:robust}. The persona-free panel receives the same question with the first sentence replaced by \pt{Rate this photograph on each of the axes below. Judge each one honestly on its own merits, and use the full range instead of defaulting to the middle:}, with \pt{artwork} in place of \pt{photograph} on LAPIS, and with \pt{in-character} removed from the comment field.
\end{sloppypar}
\begin{promptbox}{PARA question (synthetic audience)}
\ttfamily
As this annotator, rate this photograph on each of the axes below. Judge each one honestly the way this exact person would, and use the full range instead of defaulting to the middle:\\
- "aestheticScore": overall aesthetic appeal (1 = very poor, 3 = average, 5 = excellent). Rate from 1 to 5 in steps of 0.5.\\
- "qualityScore": technical image quality — sharpness, exposure, noise and artifacts (1 = very low, 5 = flawless). Rate from 1 to 5 in steps of 0.1.\\
- "compositionScore": composition — framing, balance and use of space (1 = poor, 5 = excellent). Rate from 1 to 5 in steps of 1.\\
- "colorScore": colour — palette, harmony and vibrancy (1 = poor, 5 = excellent). Rate from 1 to 5 in steps of 1.\\
- "dofScore": depth of field and focus handling (1 = poor, 5 = excellent). Rate from 1 to 5 in steps of 1.\\
- "contentScore": how interesting or compelling the subject/content is (1 = poor, 5 = excellent). Rate from 1 to 5 in steps of 1.\\
- "lightScore": lighting quality (1 = poor, 5 = excellent). Rate from 1 to 5 in steps of 1.\\
- "contentPreference": how much this person personally likes this kind of content, regardless of photographic quality (1 = strongly dislike, 5 = strongly like). Rate from 1 to 5 in steps of 1.\\
- "willingnessToShare": how willing this person would be to share this photo on their own social media (1 = never, 5 = definitely). Rate from 1 to 5 in steps of 1.\\[2pt]
Respond with ONLY a single JSON object and nothing else (no markdown, no extra text), in exactly this form:\\
\{"aestheticScore": <number>, "qualityScore": <number>, "compositionScore": <number>, "colorScore": <number>, "dofScore": <number>, "contentScore": <number>, "lightScore": <number>, "contentPreference": <number>, "willingnessToShare": <number>, "comment": "<one short in-character sentence explaining your scores>"\}
\end{promptbox}
\begin{promptbox}{EVA question (synthetic audience)}
\ttfamily
As this participant, rate this photograph on each of the axes below. Judge each one honestly the way this exact person would, and use the full range instead of defaulting to the middle:\\
- "score": overall aesthetic score (0 = very bad, 5 = average, 10 = excellent). Answer with a whole number from 0 to 10.\\
- "visual": light and colour (1 = very bad, 2 = bad, 3 = good, 4 = very good). Answer with a whole number from 1 to 4.\\
- "composition": composition and depth (1 = very bad, 2 = bad, 3 = good, 4 = very good). Answer with a whole number from 1 to 4.\\
- "quality": technical quality — sharpness, exposure, noise (1 = very bad, 2 = bad, 3 = good, 4 = very good). Answer with a whole number from 1 to 4.\\
- "semantic": semantic content — how interesting or meaningful the subject is (1 = very bad, 2 = bad, 3 = good, 4 = very good). Answer with a whole number from 1 to 4.\\
- "difficulty": how easy this image was to judge (1 = very difficult, 2 = difficult, 3 = easy, 4 = very easy). Answer with a whole number from 1 to 4.\\[2pt]
Respond with ONLY a single JSON object and nothing else (no markdown, no extra text), in exactly this form:\\
\{"score": <integer>, "visual": <integer>, "composition": <integer>, "quality": <integer>, "semantic": <integer>, "difficulty": <integer>, "comment": "<one short in-character sentence explaining your ratings>"\}
\end{promptbox}
\begin{promptbox}{LAPIS question (synthetic audience)}
\ttfamily
As this participant, rate this artwork on each of the axes below. Judge each one honestly the way this exact person would, and use the full range instead of defaulting to the middle:\\
- "rating": aesthetic appreciation — how aesthetically pleasing this person finds the artwork (0 = not at all, 50 = middling, 100 = extremely). Answer with a whole number from 0 to 100.\\[2pt]
Respond with ONLY a single JSON object and nothing else (no markdown, no extra text), in exactly this form:\\
\{"rating": <integer>, "comment": "<one short in-character sentence explaining your rating>"\}
\end{promptbox}

\subsubsection{Persona Cards for Rapidata.} Rapidata records no identifier for its voters and only two attributes, the country of each voter as a two-letter ISO 3166-1 code and their language as a two-letter ISO 639-1 code, together with a reliability score that the platform assigns. We write both codes out as English names, reading language codes without regard to case since the corpus contains both \pt{EN} and \pt{en}, and treat the placeholder codes \pt{unknown} and \pt{UD} as missing. The card then reads \pt{a person from <country> who speaks <language>}, for example \pt{a person from India who speaks English}. When only the country is known it shortens to \pt{a person from <country>}, and when only the language is known it reads like \pt{an English-speaking person}. Votes that carry no information about the voter, about half of the full corpus, are left out, so every one of the 23{,}445 votes we use names a country or a language and is judged under a card, and we did not run a persona-free panel on these pairs (Section~\ref{sec:generated}). The reliability score never enters the card, since it describes how far the platform trusts the voter rather than their taste. Without identifiers, a card describes everyone who shares a country and a language rather than one person, so two votes with the same card may or may not come from the same voter. The VLM casts one vote under the card of every human voter of a pair.

\subsubsection{Pairwise Prompt (Rapidata).} Both images go into one message, and which of them comes first is drawn at random for every vote, so a position bias toward the first image cannot be confounded with aesthetic preference. The question is the one Rapidata showed its voters, and since they never saw the caption the images were generated from, the prompt leaves it out as well.
\begin{promptbox}{Rapidata system prompt (synthetic audience)}
\ttfamily
You are role-playing as one specific participant in a worldwide, crowdsourced study of image preferences. The participant is \{description\}\\[2pt]
You are shown two AI-generated images and asked which one you prefer. Answer the way this exact person would: let their cultural background, the visual world they live in, and their own taste decide which image appeals to them more. This is personal preference, not a test — there is no correct answer, and you must pick one of the two. Do not mention that you are an AI or that you are role-playing.
\end{promptbox}
\begin{promptbox}{Pairwise question (synthetic audience)}
\ttfamily
Which image do you prefer?\\[2pt]
Answer "A" for the first image or "B" for the second. You must choose one; do not answer with both or neither.\\[2pt]
Respond with ONLY a single JSON object and nothing else (no markdown, no extra text), in exactly this form:\\
\{"choice": "A" or "B", "comment": "<one short in-character sentence saying why>"\}
\end{promptbox}

\subsubsection{Critic Prompts (Editing Loop).} The critics of Section~\ref{sec:c4} reuse the PARA system prompts above. Each of the ten personas on the panel receives the PARA system prompt of the synthetic audience with its own card, drawn once at random from the PARA raters, and the persona-free critic receives the persona-free PARA system prompt. Only the question changes, to ask for a score from 0 to 100 and the single change that would most improve the image. We record the score but never use it to accept an edit.
\begin{promptbox}{Critic question (synthetic audience)}
\ttfamily
As this exact person, look at this image. Give it an overall aesthetic score from 0 to 100 for how much YOU personally like it, then name the SINGLE change that would most improve it for your taste. Respond with ONLY a JSON object and nothing else, in exactly this form:\\[2pt]
\{"score": <integer 0-100>, "comment": "<one concrete improvement, imperative>"\}
\end{promptbox}
\begin{promptbox}{Critic question (persona-free critic)}
\ttfamily
Look at this image. Give it an overall aesthetic score from 0 to 100, then name the SINGLE most important change that would most improve it. Respond with ONLY a JSON object and nothing else, in exactly this form:\\[2pt]
\{"score": <integer 0-100>, "comment": "<one concrete improvement, imperative>"\}
\end{promptbox}

\subsubsection{From Complaints to One Instruction.} The current instruction is fed back in every round, so the complaints accumulate across rounds.
\begin{promptbox}{System prompt}
\ttfamily
You are a photo-editing assistant. You convert viewer feedback into ONE concrete, imperative image-edit instruction. Output only the instruction — no preamble, no quotes, no explanation.
\end{promptbox}
\begin{promptbox}{User prompt}
\ttfamily
[The current edit instruction is: "<accumulated>".]\\[2pt]
Viewers looking at this image want these changes:\\
- <complaint 1>\\
- <complaint 2>\\
\ \ \dots\\[2pt]
Rewrite everything into ONE concrete imperative photo-edit instruction of at most 15 words that an image editor can apply. Output only the instruction.
\end{promptbox}

\subsubsection{Editor Instructions Shared by All Critics.} The fixed-instruction baseline uses one instruction in every round, the generic-instruction cycle cycles through eight generic instructions, and the other critics write their own. For every critic, the same request for a visible edit is appended to the instruction before it reaches the image editor.
\begin{promptbox}{Fixed instruction (called the static string in the run logs)}
\ttfamily
Improve the overall aesthetic quality of this image.
\end{promptbox}
\begin{promptbox}{Generic-instruction cycle (one per round, in this order, repeating after the eighth)}
\ttfamily
Improve the overall aesthetic quality of this image.\\
Enhance lighting, contrast, and color balance.\\
Sharpen details and reduce noise.\\
Improve composition and remove distracting elements.\\
Make the colors richer and more vibrant.\\
Brighten the image and improve dynamic range.\\
Increase clarity and depth of field.\\
Give the image a cleaner, more professional look.
\end{promptbox}
\begin{promptbox}{Visible-edit request (appended for every critic)}
\ttfamily
Make this a clearly visible edit, not a subtle one, while keeping the same subject and scene.
\end{promptbox}

\section{Future Directions}
\label{sup:future}
Our results come from a single frozen VLM, Qwen2-VL-7B-Instruct, with greedy decoding of ratings, so the first extension is to repeat the same controls with sampled decoding and with other model families and sizes, to tell which of our negative results belong to persona audiences in general and which to this model. A second is richer personas, such as interviews or a few of the viewer's own ratings, since a rater's own ratings predict far more of how they depart from the crowd than their profile does (Section~\ref{sec:ceiling}). A third is interaction, since simulated viewers who see each other's reactions could either recover group structure or lose it through conformity. Finally, the editing test should be repeated with judges outside the loop and with viewers from the group the edits are meant for.

\section{Ethics}
\label{sup:ethics}
Every dataset we use was collected and released for research by other groups, and we collected no data of our own. Cards contain only attributes the datasets already publish, but rare combinations can remain distinctive, so we do not reproduce them. Biases in these records can carry into the judge, and the tilt we find for some regions and nationalities after calibration (Appendix~\ref{sup:fairness}) is something anyone deploying such a system should check on their own population. The reversed age pattern on EVA shows a further risk, since a persona audience can attribute to a group a rating pattern that runs opposite to the group's real ratings and present it as that group's taste.

\section{Reproducibility}
\label{sup:repro}
\ifdefined\arxivversion
The code, prompts, analysis scripts and derived model outputs will be released through the project page at \url{https://forall-labs.github.io/AutoPolish}, and the datasets can be obtained from their original authors.
\else
No anonymized artifact accompanies this submission, so the code, prompts, analysis scripts and derived model outputs are not yet available to reviewers. We intend to release them with the final version, and the datasets can be obtained from their original authors.
\fi No generative-model weights are updated, and every model we use is frozen and publicly available, namely \texttt{Qwen/Qwen2-VL-7B-Instruct}, \texttt{black-forest-labs/FLUX.1-Kontext-dev}, \texttt{facebook/dinov2-base} and the LAION improved aesthetic predictor on CLIP ViT-L/14. The model revisions on the Hugging Face hub and the software versions were not recorded in the run logs. Ratings use at most 128 new tokens and the editing critics 96, the merge 48. The rating runs requested a temperature of 0 (first persona run) or 0.7 (second persona run and persona-free panel) without setting top-$k$ or top-$p$, so the model's shipped generation configuration ($k{=}1$, $p{=}0.001$) made every rating argmax decoding, while the Rapidata pipeline set $k{=}0$ and $p{=}1$ and records them, so its votes are truly sampled. Answers are parsed as JSON with a regular-expression fallback, and every saved rating record has a parsed overall score, although generation-time retries were not logged. Calibration uses scikit-learn's isotonic regression with clipping, and the image halves, rater folds and bootstraps use seed 0. The group statistics, aspect comparison and editing diagnostics come from three scripts over the saved outputs. Appendix~\ref{sup:prompts} gives every prompt verbatim and names the experiment that uses it, namely the rating prompts for PARA, EVA and LAPIS, the pairwise prompt for Rapidata, and the critic and editing prompts for the editing loop, together with how the persona cards of each dataset are built. The experimental setup in the main paper describes the evaluation protocol.

\fi %

\end{document}